%% file: IEEE10.tex
\documentclass[conference]{IEEEtran}

\usepackage{diago}
\def\substform#1#2{\hbox to 4.35\goheight{#1\hskip2pt\string @\hskip2pt#2\hss}}
\def\passform#1{\hbox to 3.5\goheight{#1\hskip2ptpass\hskip2pt\hss} }

\begin{document}
\normalsize\golength11pt\gothinness0.4pt\gothickness1pt

\title{A Dynamical Systems Approach for Static Evaluation in Go}
%\runningtitle{Go Positions as Complex Systems}

\author{
\IEEEauthorblockN{Thomas Wolf}
\IEEEauthorblockA{Brock University \\
 St.\ Catharines, Ontario, Canada \\
WWW: http://www.brocku.ca/mathematics/people/wolf/ \\
Email: twolf@brocku.ca}
}

\maketitle
%\tableofcontents
\begin{abstract}
  In the paper arguments are given why the concept of static
  evaluation has the potential to be a useful extension to Monte Carlo
  tree search. A new concept of modeling static evaluation through a
  dynamical system is introduced and strengths and weaknesses are
  discussed. The general suitability of this approach is demonstrated.
\end{abstract}

\section{Motivation} \label{motivation}
The concept of Monte-Carlo simulations applied to Go
\cite{Brueg93} combined with the UCT algorithm \cite{ACF02,KS06},
which is a tree search method based on Upper Confidence Bounds (UCB) 
(see e.g.\ \cite{Bouzy03monte-carlogo})
produced a new type of programs like \cite{MoGoBot,MoGoBot2,CrazyS} 
%,GS07,Bouzy03monte-carlogo,Gelly2006 
that dominate computer Go in recent
years. The detailed tournament report \cite{Taiw09} of the program MoGo
playing against professional and amateur players reveals strengths and
weaknesses of MoGo which are typical for programs that perform a 
Monte Carlo tree search (MCTS). 

%Based on an early concept for Monte-Carlo (MC) simulations applied to Go
%\cite{Brueg93}, this approach started to dominate computer Go in recent
%years when combined with the UCT algorithm \cite{ACF02,KS06},
%which is a tree search method based on Upper Confidence Bounds (UCB).
%This type of approach (see e.g.\ \cite{Bouzy03monte-carlogo})
%produced new programs -- \cite{MoGoBot,MFoG,CrazyS} --
%%,GS07,Bouzy03monte-carlogo,Gelly2006
%that are stronger than the previous best programs, by an equivalent of
%5-6 handicap stones. For the first time it was possible to beat high
%dan professional players (at least in the first game) when starting
%with 7 handicap stones (\cite{NWO09}).

Programs performing MCTS can utilize ever increasing computing power
but in their pure form without extra Go knowledge the ratio
log(increase in needed computing power) / (increase in strength) is
too big to get to professional strength on large boards in the
foreseeable future.  Therefore in recent years Go knowledge has been
incorporated either in form of heuristics, or pattern databases
learned from professional games or from self-play. Although treesearch
was naturally slowed down the playing strength increased further.

With all of this tremendous progress of MCTS compared to the knowledge based
era of computer Go summarized in \cite{KC01a}, \cite{BC03}, \cite{MM02b}, 
%especially with the progress in $9\times 9$ Go where programs like
%MoGo running on large hardware reached professional level,
it needs good reasons to start work on a static evaluation function
(SE)
%\footnote{Among users of the Internet Go server KGS the
%abbreviation SE is used for 'Score Estimator'. Although different from
%'Static Evaluation' a score estimator is easily obtained from static
%evaluation by adding up probabilities of blocks to be alive at the end
%of the game or points to be owned by White or Black.}  
in Go.

One indicator that more Go knowledge needs to be added is that,
compared with human playing strength the playing level of current
programs decreases as board size increases from $9\times 9$ to
$13\times 13$ and then to $19\times 19$.

%\item There are still children under 10 years old with a clearly
%  higher playing strength than the best programs running on the
%  largest clusters available. Even if computing power could be
%  increased by currently unimaginable amounts and minors could be
%  beaten in the game, it would be unsatisfying, if playing strength 
%  would have to rely so massively on hardware.
%\end{itemize}

The principal difficulties of deriving knowledge and applying it become more
relevant as knowledge is increasingly used in MCTS. 
\begin{itemize}
\item Knowledge that is not 100\% accurate reduces the scalability of
the program when enough computing power is available for global search
to replace increasingly the approximate Go knowledge which then becomes 
less useful or even less accurate than knowledge coming from search.
\item It is difficult to combine knowledge on a high level if it comes
from different sources, like from pattern and from local searches. It
is one of the reasons of the originally surprising success of pure MCTS
that it only uses knowledge from one source (statistics of
simulations) without the need of merging different types of knowledge.
\item Go knowledge is either accurate, for example, when coming from
life and death computations of fully enclosed regions or from matching
large patterns but then these pure situations occur only rarely, or, Go
knowledge is produced more often, for example, by simply counting liberties
or matching small
pattern or by doing quick life and death computations in open regions but
then the knowledge has little significance or is not very accurate.
\end{itemize}

%One way to equip MCTS with more Go knowledge is to add knowledge
%directly by having heuristic procedures detect special situations and
%add a bonus to the number of simulations won by moves that it
%recommends.
%
%The problem with this approach seems to be that 
%\begin{itemize}
%\item either the situations that allow precise recommendations of
%  moves are too special and thus happen too rarely, so that the
%  constant effort invested in recognizing such positions is too
%  high compared with the improvement MCTS would gain given the same 
%  extra time, or
%\item the situations are more general, but this generality prevents an
%  accurate recommendation of the best move or a safe recommendation of which
%  moves to ignore, or
%\item anything in between.
%\end{itemize}

What is needed is an approach that has an understanding of the local and
global situation in {\em general} positions, that either is accurate to some
extent or indicates in which areas it is not accurate. On top of that it
needs to be fast. 
% What simplifies matters is that it does not have to be
% perfect in any respect. Lack of accuracy could be compensated with a higher
% speed, or it could be slow if its accuracy matches the quality of the
% intuition of a strong human player.

The SE proposed in this paper is applicable to any position, it is
reasonably fast and has already in its first and simplest version as
studied in this paper a good understanding of the situation. It is
expandable to include linking and life and death information in future
without having to combine knowledge in a simplistic way, for example,
by taking an average with constant weights. 

%\section{Initial Considerations} \label{initial}
%  \subsection{Resources Unused in Pure MCTS}
%  \subsection{Design Decisions}
In section \ref{initial} a close look at characteristic
features of the game Go provides arguments for
using a dynamical systems approach as the starting point for a 
static evaluation function.
%\section{About Discrete Dynamical Systems} \label{DDS}
%  \subsection{General Comments}
%  \subsection{Potential Benefits of Dynamical Systems}
%  \subsection{Computational Complexity}
Section \ref{DDS} gives an introduction to discrete dynamical systems 
and a description of their potential benefits.
%\section{A Dynamical System Representing a Board Position} \label{dynsys}
%  \subsection{The Setup}
%  \subsection{State Variables}
%  \subsection{The Relations}
%  \subsection{An Example Computation}
%  \subsection{A Full Board Example}
The algorithmic details of formulating, initializing and evaluating
dynamical systems are described in section \ref{dynsys}. 
%\section{Limits of what Dynamical Systems can do} \label{cannot}
%  \subsection{Ladders}
%  \subsection{Life \& Death}
Section \ref{cannot} comments on limitations of any dynamical 
systems approach that are based on its static nature.
%\section{Results} \label{results}
%  \subsection{Existence, Uniqueness and Stability}
%  \subsection{Statistics on Professional Games} \label{stat}
%  \subsection{Interpretation}
%  \subsection{Timing}
%  \subsection{Parameter Dependence}
Section \ref{results} discusses the robustness of the approach
(existence, uniqueness and stability of fixed points) and its
efficiency in predicting moves in professional games.
%\section{Comparisons with other static evaluations} \label{comparisons}
%  \subsection{Criteria for Static Evaluations}
%  \subsection{Comparisons}
%  \subsection{Characterization of SEDS}
A comparison with the efficiency of other approaches 
is made in section \ref{comparisons}. This section also
contains a detailed characterization of
static evaluation through dynamical systems.
%\section{Future Tasks} \label{future}
%  \subsection{Improvements of the Algorithms} \label{improvements}
%  \subsection{Multipole Moments of Influence} \label{multipole}
%  \subsection{Merging Static Evaluation and Treesearch}
Section \ref{future} includes comments on necessary extensions, 
especially the merger with MCTS, but also comments
on how an appropriate influence function can be used in a game.
%\section{Summary} \label{summary}
%\section*{Appendix}
The paper concludes with a short summary in section \ref{summary} and
an appendix that gives an example how an influence function can guide
finding the best move even in situations when a sacrifice move is
necessary.

%Before supporting these statements with test details in section \ref{results},
%we argue in the following section why a high quality SE is not unrealistic and
%how it should be structured. We give details of the dynamical system in
%section \ref{dynsys} and discuss limits of the design in section \ref{cannot}.
%In section \ref{comparisons} the new approach is compared with other
%evaluation functions and its strengths but also necessary future improvements
%are described. A short summary follows in section \ref{summary}. The appendix
%gives an example how an influence function can guide finding the best move
%even in situations when a sacrifice move is necessary.

\section{Initial Considerations} \label{initial}
\subsection{Resources Unused in Pure MCTS}
Why should it be possible to design a static evaluation which can 
provide at least some information faster and/or better than MCTS?

One strength of MCTS is to be useful for programs playing other games
than Go or even for work on tasks not involving games. This
strength is at the same time a weakness when applied to Go: 
MCTS does not take advantage of simplifying aspects of the nature of Go:
\begin{enumerate}
\item {\em Blocks are local.} Blocks connect adjacent stones of the same
  colour into a unit, so that either all of them are captured or none of
  them is. Stones that are not connected to the block do not belong
  to it.
\item {\em Capturing a block is local.} To capture a block the opponent must
  fill all of the block's {\em adjacent} intersections. Opposing stones
  placed further away do not capture the block.
%\item Neighouring blocks may influence each others strength gradually.
%  For example, a black block may be strong because some of its
%  neighbouring white blocks are weak, because all their neighbouring
%  black blocks are strong.
%\item {\em go has an integer score.}  In contrast to Chess or Checkers
%  which have just a one bit outcome, in go the result is an integer,
%  which is a first indication that there exist some measure (value,
%  influence, strength, aji) which might be useful as a model to guide
%  towards optimal play.
\item {\em Go has an influence field.}  
  At all stages of a game except its very end it is useful to
  introduce a field of 'influence' or 'strength' that guides the
  player (humans and computer) towards optimal play.

  This influence field is not simply a tool to accommodate human slowness in
  reading compared to MCTS. The point to make is that this
  influence field is in some sense real and can be characterized and
  modeled. It shows, for example, some stability or null-sum property and can
  explain higher playing level sacrifice moves as done with a position in the
  appendix.

  We want to state it as a conjecture: {\em To maximize playing
  strength for a given amount of computational power (size of memory
  and cycles per second, both sufficiently large but fixed) a field
  embodying strength or influence and perhaps other fields or state
  variables have to be introduced.} This is not different from
  progress in the natural sciences and mathematics where the
  improvement of quantitative knowledge and an evolution of the
  scientific language depend on each other.

  Knowledge on strength and influence is crucial for human players but
  also increasingly used in MCTS. Deriving knowledge purely from
  search as in original MCTS results in a search space growing
  exponentially with the area of the board. In contrast, the cost of
  computing a good static evaluation does not increase exponentially
  with the board size.

\item {\em Influence varies smoothly.} The influence of stones falls off
  smoothly, at least in the opening in non-life-and-death situations.  Also,
  the example in the appendix shows the need for intermediate influence values
  other than 1 (full domination) and 0 (no influence at all).  More discussion
  on smoothness in Go can be found in section II.A.\ in \cite{WolfIEEE09}.

%  At higher playing level Go shows features of a null-sum game.
%  Let us assume there is an extended area on the board, $A$, which is
%  weakly dominated by, say, White with Black still having aji
%  (potential) in this area, but not enough influence to live anywhere
%  in $A$ if Black would only play locally in that part of $A$. Then
%  Black may still have the possibility to play one or more sacrifice
%  stones and as a result force White to get total dominance around the
%  sacrificed stones in exchange for Black bundling enough of this
%  previously scattered aji to get permanent presence in another part
%  of $A$. (A more detailed discussion of smoothness of Go is given in
%  section II.A.\ in \cite{WolfIEEE09}.)
\end{enumerate}

\subsection{Design Decisions}
The above observations lead to the following design decisions 
for the static evaluation function.
\begin{itemize}
\item We want to start by modeling/computing only a minimal number of
  variables describing a board position. This will be for each point
  ({\em empty} intersection) a number indicating whether it is under
  Black or White influence and for each block a measure of its
  strength, i.e.\ a probability of not being captured.
\item The strength values of blocks and influence values at
  points are represented by floating point numbers because of 
  points 3) and  4) above but also to evaluate some fuzzy knowledge 
  by a number that changes smoothly with the degree of certainty of the
  knowledge.
\item Because of points 1) and 2) above, the set of all relations of
  neighbouring points and blocks is formulated as a single discrete
  dynamical system of algebraic relations expressing the strength of
  each block and influence at a point in terms of the strength of
  neighbouring blocks and influence at neighbouring points. A more
  detailed description of dynamical systems is given in the subsection
  below. In the approach to be described a strength value is assigned
  to a whole block, irrespective of its size or shape. All that
  matters in this approximation are the neighbourhood relations.  A
  {\bf S}tatic {\bf E}valuation based on {\bf D}ynamical {\bf S}ystems
  will be abbreviated as SEDS in the remainder of the paper.
\end{itemize}
A different question is whether the SE should depend on who moves
next. Although it may become slightly better by taking that into
account (e.g.\ if two important blocks of opposite colour touch each
other and have only one liberty each, so that the side to move next 
may capture the opponent's block), the SE to be described in this 
contribution does not use who moves next. The reasons for this are:
\begin{itemize}
\item It is not obvious how to use who moves next without 
  prejudice even in the
  simple case of, say, Black moving next and white blocks under
  atari being so small that their capture has low priority. Another
  example is the case when many white blocks are under atari.
\item Making the SE dependent on who moves next is not a general
  solution. It may take 2 moves to simplify the all-or-nothing fight
  so that the SE can 'see' the outcome, or 3, 4, ... moves. The issue
  of merging SE and MCTS has to be solved more rigorously, not by a
  quick fix of making SE dependent on who moves next.
\item The value of moving next naturally varies from area to area. To
  consider it properly would imply to know the value for each area but
  that essentially means to be able to play perfectly. This would be
  contradictory to the philosophy of splitting up the problem of
  determining the best move into three parts: designing a static
  evaluation, a search procedure (MCTS) and an interplay between both.
\end{itemize}

\section{About Discrete Dynamical Systems} \label{DDS}
\subsection{General Comments}
A discrete dynamical system (abbreviated as dynamical system (DS) in
the following) is a set of $n$ relations between $n$ variables where
each relation takes the form of expressing one variable in terms of
all others:
\begin{equation}
v_i = f_i(v_j) \label{DS}
\end{equation} 
where $f$ is some (not necessarily continuous) map.\footnote{If $v_i$
would be functions of a parameter, e.g.\ time, then dynamical systems
typically express the time derivative of each variable in terms of all
variables (not their time derivatives).
% By describing the dynamics (time-dependence) of a set of variables
% such systems are called 'dynamical systems'.
}

Given numerical initial values $v_i=v_{i0}$ the system of relations
can be used to compute new values $v_{i1}, v_{i2},..$ until after $k$
iterations a fixed point is reached, i.e.\ $v_i$ change only 
insignificantly: 
      \[|v_{ik}-v_{i(k-1)}| \leq \varepsilon \ \ \ \forall i\] 
for some threshold parameter $\varepsilon$, or, until $k$ reached an
upper bound $k_{\rm max}$. 

In our approach the state variables $v_i$ are strength values of blocks
and influence values at points (empty intersections). By choosing
suitable maps $f_i$ the iteration of the system (\ref{DS}) should
model, for example, the fact that blocks are weakened when being under
strong opponent influence and as a consequence in the next iteration
will become less influential on their surrounding.

%Capturing essential aspects 
A dynamical system is called {\em sparse} if the $f_i$ involve only
few variables. The DSs we aim at are sparse due to locality properties
1), 2) above: each empty point has at most 4 neighbours (points or
blocks) and each block has only a small subset of all points and
blocks as direct neighbours. If a DS is sparse then iterations are
computed faster than if it would be {\em dense}, i.e.\ if many $f_i$ 
would involve many $v_j$.

Although only local relations are recorded in the formulation of a DS,
the fixed points that are computed are global properties of the
complete DS due to the iterations that take place.

\subsection{Potential Benefits of Dynamical Systems}
The potential benefits of using DS to describe a board position in Go
are manifold.

\begin{itemize}

\item It allows a consistent formalism where the influence originating
  from a block depends on the strength of the block and the strength
  of the block depends on the degree of ownership it has on
  neighbouring empty points. 

  In other computer Go programs which use influence functions or
  pattern matching either all blocks have the same strength or are
  distinguished only between being alive or dead. The error made with
  this crude simplification needs to be compensated by global search,
  i.e.\ by giving the influence function and pattern heuristic
  relatively little weight compared to the weight of search. Whether
  supposedly better influence and strength values computed from
  finding fixed points of a DS are worth the extra effort remains to
  be shown but the potential is at least there.

\item Data about which points and blocks are neighbour to each other
  are recorded to speed up the computation of $f_i$ in (\ref{DS}).
  These data are available for other computations, like
  life and death.

\item The strength value of blocks and influence values at points
  become available as a side product which may be useful for separate
  tactical investigations. Especially pattern matching is very popular
  among computer Go programmers since the very start of computer
  Go. Having adequate strength and influence values may allow more
  refined pattern which encapsulate not just local data but global
  board properties because the numerical strength and influence values
  used in the local pattern are properties of the computed fixed
  point, i.e.\ their values take into consideration the complete
  board. As another example, strength values of blocks may be useful
  to initialize the area of local life \& death computations.

\item The number of iterations needed for $|\Delta v_i|$ to fall below
  $\varepsilon$ turns out to be a good measure for the stability of a
  local region on the board and thus a strong indicator for which
  moves need to be searched, e.g.\ through (some local version of) MCTS.

\end{itemize}

\subsection{Computational Complexity}
For a programmer of MCTS
the computational effort to formulate a DS and find a fixed point may
be horrifying.
%seem completely and utterly too high. 
After-all, in MCTS tens of thousands of whole games starting from a
position are played just to find the next move. Speed in performing
moves is essential in this approach, even to a point that it may pay
off to allow illegal moves in order to save time and play more
simulated games.

In total contrast a dynamical system has several hundred variables
and as many equations (
% = the number of blocks + 2 times the number of empty points = 
489 variables and relations for the full board position in diagram
\ref{FullBoard} further below). At first the DS has to be established in some
form, i.e.\ enough data need to be collected and stored to compute the
$f_i$ in (\ref{DS}) efficiently. Then this system has to be iterated
repeatedly to compute a fixed point. Next, an estimated score is to be
computed by summing over all blocks their product (strength $\times$
size) and summing over the influence values of all empty points. All
of this computation has to be done after performing each one of the
legal moves that are to be evaluated to select the move with the
highest score.

But, {\em if} local relations 1), 2) are an important part of the
game, and {\em if} the strength of blocks and the influence at empty
points are inherent characteristics of a board position then a DS
(\ref{DS}) is the most direct and adequate formulation of all
relationships on the board and solving such a systems (finding the
fixed points) is the most efficient way to characterize the board
position. Other methods will have at least as high costs whether
obvious or hidden.

These are a lot of {\em if}'s which are not always satisfied. For
example, in the case of a lengthy winding ladder there is no
alternative to performing the tree search and computing the
ladder. But many parts of the board especially in the opening are calm
where the {\em if}'s are satisfied and where the formulation and
solution of a DS appears to be the appropriate computation.

Time measurements shown at the end of section \ref{results} and
comparisons with other evaluation functions in section
\ref{comparisons} indeed show that the DS-approach is fast.

Another useful feature is that the evaluation of different moves based
on their score according to SEDS can be done in parallel. Especially
when more life \& death investigation will be added to SEDS to make it
more accurate, the parallelization will become more coarse grain and
thus be more effective.

%Work on making SEDS thread safe (for performing a parallelization 
%through threats) has been started but is not completed yet.

%---------------

\section{A Dynamical System Representing a Board Position} \label{dynsys}

In this section we describe a conceptually simple dynamical system
model that was implemented and studied for its strengths and
weaknesses.

\subsection{The Setup}
The elementary objects on the board (we call them units from now on)
are taken to be all points (empty intersections) and blocks (for which
no shape is recorded).  Individual stones of a block have no own
identity in this model.

Based on the capture rule of Go, units have completely local relations
with each other, i.e.\ the state variables describing each unit can be
computed explicitly from the state variables of neighbouring units and
the resulting dynamical system can be solved iteratively.

This system couples all units on the board (i.e. all (empty) points
and blocks) and thus a fixed point of the dynamical system is a global
consequence of the whole board. A change in strength of one block
would influence the strength of weak neighbouring blocks and so on but
the change of influence would stop at strong blocks.

With points on the edge of the board having only 3 neighbours and in
the corners only having 2 neighbours, the influence of the edge should
come out properly without the need of extra artificial adjustments.

\subsection{State Variables}
To each {\em point} $i$ (i.e.\ each empty intersection, i.e.\ $i$
takes at most $19\times 19$ different values) 
are attached two real floating point type numbers
describing probabilities at the end of the \vspace{9pt} game:\\ 
$w_i\ldots$ to be owned by \twpstone\ , i.e.\ 
to be occupied by \twpstone\ or to be a point in an alive white eye\\
$b_i\,\ldots$ to be occupied by \tbpstone\ , i.e.\ 
to be occupied by \tbpstone\ or to be a point in an alive black 
\vspace{9pt} eye\\
and to each {\em block} $j$ is attached one number: \vspace{9pt}\\
$s_j\ldots$ probability for this block \vspace{9pt} to survive.\\ 
All values are in the interval $0\ldots 1$. \\
For explanation purposes we also  \vspace{9pt}introduce\\
$\bar{w}_i,\ \bar{b}_i\ldots$ probability that at least one
neighbouring intersection of point $i$ is occupied by 
resp.\ \twpstone\ or \tbpstone\ at the end of the game.

\subsection{The Relations}
If $b_i, w_i$ are the probabilities defined above then under normal
circumstances they add up to 1:
\begin{equation} 
b_i+w_i=1 .   \label{e1}
\end{equation}
The only exception is a seki, for example, when point $i$ is one of
the shared liberties that is not accessible to either one side. In
that case we should have $b_i=w_i=0$ but our model will give
$b_i=w_i=0.5$ which does not change the contribution of 
points to the score but it may have an effect on the computed
strengths $s_j$ of the blocks in seki.  The SEDS could conclude from
$w_i=b_i=0$ that both blocks are safe, but not from $w_i=b_i=0.5$. The
bottom line is that the concept of seki like that of life \& death, is
a discrete concept resulting from the rule that both sides alternate
moves. These concepts are not properly covered in this dynamical
systems approach but have to be detected separately, see the
discussion in section \ref{cannot}.

Apart from relation (\ref{e1}) the only other
assumption we make is \ \ $w_i/b_i = \bar{w}_i/\bar{b}_i$\ , \ \ i.e.\ 
\begin{equation} 
w_i\bar{b}_i = b_i\bar{w}_i .  \label{e2}
\end{equation}
At least in the extreme cases 
$(\bar{w}_i,\bar{b}_i)=(1,1),\ \ (1,0),\ \ (0,1)$ 
this relation is correct. Also for other cases it should be suitable 
based on the following argument.

Under Chinese rules the aim is to occupy as many points as possible at
the end of the game. A single stone can not be alive on its own if
surrounded only by alive opponent stones, it has to be part of a block
that is alive. Thus, the probability $b_i$ of a point being owned by 
Black should increase with the probability that it has an
alive black stone as neighbour, i.e.\ $b_i \propto \bar{b}_i$ and
similarly $w_i \propto \bar{w}_i$. Relation (\ref{e2}) is a simple
example with such a dependence.

From (\ref{e1}), (\ref{e2}) we get
\[w_i=\frac{\bar{w}_i}{\bar{b}_i}\big(1-w_i\big)=\frac{\bar{w}_i}{\bar{b}_i}
                                       -\frac{\bar{w}_i}{\bar{b}_i}w_i\]
\[w_i\left(1+\frac{\bar{w}_i}{\bar{b}_i}\right)=\frac{\bar{w}_i}{\bar{b}_i}\]
\begin{equation}
  w_i=\frac{\bar{w}_i}{\bar{b}_i} 
      \left(1+\frac{\bar{w}_i}{\bar{b}_i}\right)^{-1}=
      \frac{\bar{w}_i}{\bar{w}_i+\bar{b}_i} \label{e3}
\end{equation}
where $\bar{w}_i,\bar{b}_i$ have to be expressed in terms of
$b_j,w_j,s_j$ from the neighbouring points and blocks. What formula
(\ref{e3}) achieves is to express the influence in a point in terms of
the status at direct neighbouring intersections.

\subsection{An Example Computation}
\noindent\parbox{0.29\textwidth}{ In this simple example we are going
to use relation (\ref{e3}) to compute the probability of points 1, 2, and 3
in diagram \ref{dia3} to be occupied finally by \tbpstone\ or \twpstone\ . 
For the simplicity of this example, all blocks are set to be alive:
$\,s_j=1,\ \forall j$. In the real model}
\hspace{0.03\textwidth}
\parbox{0.15\textwidth}{
\[ \myblock {0.15\textwidth} {
  \board4--10/5--9/
  \goboardini\alphafalse
  \whites{F6G6H6F8G8H8}
  \blacks{E7J7}
  \purenumbers[1]{F7}
  \purenumbers[2]{G7}
  \purenumbers[3]{H7}
  \shipoutgoboard
 }{}{}{dia3}\]\vspace*{3pt}
} 
their strengths would also be computed iteratively based on the strength 
of direct neighbouring blocks and the influence of their direct
neighbouring points.

To apply (\ref{e3}) we set $\bar{b}_1=s({\rm left\ black\ stone})=1$ and
$\bar{w}_1=s({\rm white\ stones})=1$ and get 
\[\rightarrow \  w_1=\frac{1}{1+1}=\frac{1}{2}=b_1=w_3=b_3\ \ 
  {\rm (by\ symmetry)}\vspace*{-6pt}\]
\parbox{0.127\textwidth}{
\begin{eqnarray*}
\!w_2&\!\!=\!\!&\frac{1}{1+\bar{b}_2} \\
     &\!\!=\!\!&\frac{1}{1+3/4} \\
     &\!\!=\!\!&\frac{4}{7} \\
\!b_2&\!\!=\!\!&\frac{3}{7}
\end{eqnarray*}
}%\hspace{0.1\textwidth}
\parbox{0.35\textwidth}{
\begin{eqnarray*}
  \rightarrow
  \bar{b}_2&\!\!=\!\!&{\rm probability\ of\ \tbpstone\ on\ 1\ or\ 3} \\
           &\!\!=\!\!&1-{\rm probability\ of\ \twpstone\ on\ 1\ or\ 3} \\
           &\!\!=\!\!&1-w_1 w_3 \\
           &\!\!=\!\!&1-\frac{1}{4}\\
  \nwarrow &\!\!=\!\!&\frac{3}{4}
\end{eqnarray*}
}\vspace*{-6pt}

This small example demonstrates how the computation goes but it also
shows the limited value of the numbers obtained. They make sense if
the moves are played randomly. In the derivation of the formulae all
moves are assumed to be uncorrelated, but that is not the case: if
White plays on 1 then Black plays on 3 and vice-versa (if there is
nothing more urgent on the board).

A similar computation is done for all blocks where the probability of
a block $j$ being captured $( = 1.0-s_j)$ is computed as the
probability of all attached opponent blocks $k$ being alive and
all neighbouring points $i$ being occupied/dominated by the opponent:
\begin{equation}
s_j=1.0 - \prod_k s_k 
          \times
          \prod_i \left\{\begin{array}{ll}
                            w_i & {\rm if\ block}\ j\ {\rm is\ black} \\
                            b_i & {\rm if\ block}\ j\ {\rm is\ white}
                         \end{array}
                  \right.                 \label{e4}
\end{equation}
Because the move taking the last liberty of a block can not be
suicide, formula (\ref{e4}) is modified slightly. The lowest value of
the $w_i$ (resp.\ $b_i$) is increased, in the simplest choice to 1.0.
In formula (\ref{e4}) we again, for simplicity, assume non-correlation
of the feasibility of the opponent capturing moves which strictly
speaking is not justified.

%Another systematic error is made in relation (\ref{e1}). There are
%points where $w_i=b_i=0$. For example, in a living white eye there is
%$w_i=1$ although White will not play there but that is no problem if
%points with $w_i=1$ are regarded as either occupied in future {\em or}
%owned by White. But points where relation (\ref{e1}) is truly violated
%are liberties that are shared by two blocks in seki. Here is
%$w_i=b_i=0$. The problem is not the territorial count when both points
%get $w_i=b_i=0.5$ instead of $0.0$.  A problem may arise with the
%safety of the blocks.  The SEDS could conclude from $w_i=b_i=0$ that
%both blocks are safe, but not from $w_i=b_i=0.5$.

\subsection{A Full Board Example}
On the board in diagram \ref{FullBoard} are 55 blocks and 217 empty
points giving a system of $2\times 217 + 55 = 489$ equations for the
217 $w_i$, 217 $b_i$ and 55 $s_j$ variables. Effectively the problem
involves 217+55=271 variables because of $w_i+b_i=1$.

The following are just three of the 489 relations:
\begin{eqnarray}
   w_{r8}&=&(b_{q8}b_{r9}s_{r7}s_{s7} - b_{q8}b_{r9}s_{r7} + 1)/ \nonumber \\
         & &(b_{q8}b_{r9}s_{r7}s_{s7} - b_{q8}b_{r9}s_{r7} + \nonumber \\
         & & \ s_{r7}s_{s7}w_{q8}w_{r9} - s_{s7}w_{q8}w_{r9} + 2), \nonumber \\
   b_{r8}&=& - w_{r8} + 1, \nonumber \\
   s_{r7}&=& - s_{r4}s_{s7}w_{p7}w_{q6}w_{q8}w_{r8} + 1. \label{sr7}
\end{eqnarray}
The full set is shown on {\tt \small http://lie.math.brocku.ca/
twolf/papers/WoSE2010/1}. Through this system each dynamical variable
(2 for each point, 1 for each block) is expressed in terms of the
variables describing their neighbouring points and blocks.
\[
\myblock {0.45\textwidth} {   % <-- new
\board1--19/1--19/
\goboardini
\alphatrue
\board1--19/1--19/
\whites{D12D13D14C13D15E15F15F14G15G16G17M15M16N15O15O14O13C10C11D10B11
        E10F10F11G11H18J18K18K17L18B4B14B15C4D2D3D5D7D17E4E5E7E17E18E19
        F18G9H3H5H6J7J13K13L13L14M11M13N6P6P16P17P18Q5Q13Q14R4R5R6R14S7S8}
\blacks{A15B3B12B13B16B18C2C5C7C12C14C15C16C17C19D1D11D16D18E2E11E12E13
        E16F5F7F12F16F17G2G3G5G7G10G12G13G14H8H10H11H12H15J8J14J16J17K3
        K12L15L16L17M9M14M17M18N14N17O4P4P13P14Q4Q7Q12Q17R3R7R12R13R15R16S3S14}
\shipoutgoboard
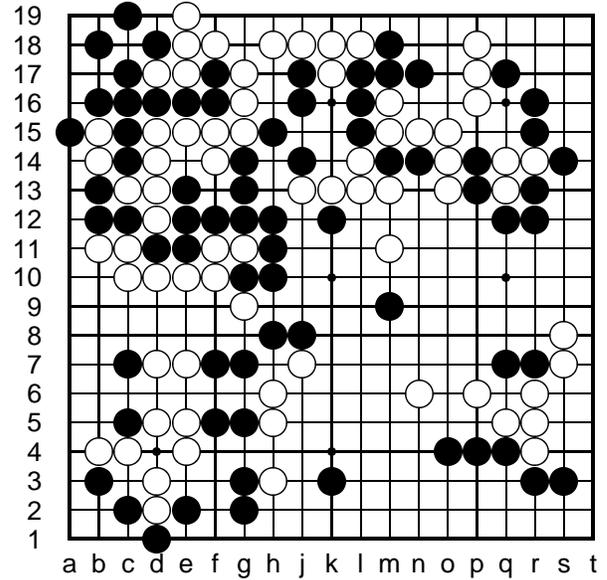
}{}{A full board position represented by a \\ 
\mbox{\ \ \ \ \ \ \ \ \ \ \ \ \ \ \ \ \ } dynamical system.}
{FullBoard} % <-- new
\]
For example, the probability of the block with the stone on $r7$
to be captured is $1.0-s_{r7}$ and is equal to the probability of 
the blocks with stones at $r4,s7$ not to be captured $(=s_{r4}s_{s7})$ and
the points $p7,q6,q8,r8$ being occupied by White $(=w_{p7}w_{q6}w_{q8}w_{r8})$ 
giving relation (\ref{sr7}).

Before the iteration all $b_i$ and $w_i$ variables are initialized to
0.5 and all $s_j$ variables are initialized to 1.0 . Then the system
is iterated until all values change less than some threshold parameter
(for example $10^{-5}$, its precise value is not crucial) at most 
some maximal number $k_{\rm max}$ of times (measurements
reported in section \ref{stat} use $k_{\rm max}=5$). 

%----------------------- 
\section{Limits of what Dynamical Systems can do} \label{cannot}
Not all rules of Go are local by nature. The rule that players
alternate in their moves sets limits to the usability of SEDS which
are to be discussed in this section. On the other hand, in a local
fight both sides may not alternate their moves.  If the fight does not
have highest priority then one side may not answer an opponent's move
and play elsewhere. A different example of non-alternating moves is
given in the appendix where a sacrifice move allows Black to move 
afterwords twice in a row in a crucial area.

\subsection{Ladders} 
The action at a distance inflicted by ladder breaking stones seems to
be a good example against local models like our dynamical system
model. But as with plane waves in physics, with long distance effect
being described through {\em local} differential equations one can not
easily dismiss the possibility that ladder breaking stones could be
described through a local model. However, ladders are a good
counterexample against static evaluation. For example, to work out a
long winding ladder like in diagram \ref{dia4} statically - even if
possible at all in principle - would be so much more difficult than
simply performing the moves in a deep but narrow tree search.
  \[ \myblock 
     {0.4\textwidth} 
     {\goboardini
      \alphafalse
\blackfirsttrue
\board1--19/1--19/
\whitecrosses{R12R13R14Q14Q15}
\whites{B3C4E5E7F1F10F11G6H7H15J7K3K11K12M12M14O15P8Q10S9}
\blacks{D1D4E6F5F6G5G11H5H6H11J3J8J11K7K13L12L13M9M15N11P14Q9Q12Q13
        Q16R9R15S11S12S13S14}
\numbers{P15R11R10Q11P10P11O11P12P13O12N12O13O14N13N14M13L14M11L11M10
         N10L10K10L9L8K9J9J10K8K10J12H10G10H9H8G9F9G8F8G7F7J6K6J5K5J4
         K4H4H3G4G3F4E4F3F2E3D3E2D2G2E1F2H2}
\shipoutgoboard
     }{}{\ \tbpstone\ catches \twcrstone\ in a winding ladder.}{dia4}  
\]

\subsection{Life \& Death}
An example for a concept in Go that is {\em not} a purely
local phenomenon, i.e.\ it can not be described by considering only
one block/point and its neighbours at a time, is the concept of {\em
  life} which is defined recursively: {\em A block is alive if it
  participates in at least two living eyes and an eye is alive if it
  is surrounded only by living blocks.}  To identify unconditional
life one has to consider the complete group of living blocks at once.

\noindent
\[
\parbox{0.2\textwidth}{
  \[ \myblock 
     {0.15\textwidth} 
     { \goboardini
       \alphafalse
       \board1--9/11--19/
       \whites{A14B14B15C14D14E14F14F15F16F17F18F19E18A16
               A17A18B19C19D19}
       \blacks{A13B13B16B17B18C13C15C16C18D13D15D17D18E13
               E15E16E17F13G13G14G15G16G17G18G19}
       \shipoutgoboard
     }{}{}{dia1}  % \vspace*{3pt}
  \]
}
\hspace{0.05\textwidth}
\parbox{0.2\textwidth}{
  \[ \myblock 
     {0.15\textwidth} 
     { \goboardini
       \alphafalse
       \board1--9/11--19/
       \whites{A18B17B19C16C18D15D17E14E16F15F13G13G14}
       \blacks{A16A17B15B16C14C15C19D13D14D18D19E12E13E17
               E18F12F16F17G12G16H12H13H14H15H16}
       \letters{G15}
       \shipoutgoboard
     }{}{}{dia2}  % \vspace*{3pt}
  \]
}
\]

For example, the life of the white blocks in diagram \ref{dia1}
depend on each other and the conclusion that all blocks are alive can
only be drawn at once, not in an iterative way and not by considering
only one block and its neighbours at a time, so not by a purely local
algorithm.  Similarly, the life of the white stones in diagram
\ref{dia2} hangs on who moves next at the {\em distant} point A in a
{\em discrete, non-iterative} way.

The current version of a SEDS computer program recognizes (non-local)
static life (life without ever having to answer any threat) at the
time when neighbourhood relations are established during the
initialization of SEDS.  Although this is a first step towards
including life and death in SEDS, static life happens only rarely
in games.

\section{Results} \label{results}

\subsection{Existence, Uniqueness and Stability}
The dynamical systems formulated along the lines of the previous
section always have at least two solutions: one solution where all
white blocks live, all black are dead and all points are fully
under white influence and the same with switched colours. If all $w_i,
b_i, s_j$ are initialized according to one of these solutions, the
iteration will keep these values stable. 

In addition to these solutions in any board position computed so far
the dynamical system had exactly one other solution (i.e.\ a fixed
point of the dynamical system) with all values in the interval
0..1. This solution was obtained from any initial conditions other
than the ones leading to the two extreme solutions mentioned
above. \vspace*{3pt}

\noindent\parbox{0.2\textwidth}{\ \ For example, when extending the
position of diagram \ref{dia3} to the one in diagram \ref{dia3a},
using just for simplification the symmetry to identify points 2 and 4 
and points 1 and 5,}
\hspace{0.001\textwidth}
\parbox{0.27\textwidth}{
\[ \myblock {0.15\textwidth} {
  \board2--10/5--9/
  \goboardini\alphafalse
  \whites{D6E6F6G6H6D8E8F8G8H8}
  \blacks{C7J7}
  \purenumbers[1]{D7}
  \purenumbers[2]{E7}
  \purenumbers[3]{F7}
  \purenumbers[4]{G7}
  \purenumbers[5]{H7}
  \shipoutgoboard
 }{}{}{dia3a}\]\vspace*{3pt}
} 
assuming that all stones are alive (i.e.\
$\bar{b}_1=\bar{w}_1=\bar{w}_2=\bar{w}_3=1$) and using formula
(\ref{e3}) repeatedly then the system of equations can be boiled down
to one equation of degree 3 for $w_2$: $0=4w_2^{\ 3} - 2w_2^{\ 2} - 7w_2 + 4$
which has numerical solutions 1.259, 0.589, -1.348, only one of
which is in the interval 0..1. But that is no lucky coincidence.
Equation (\ref{e3}) guarantees that the single value for $w_i$
lies in the interval $0 \leq w_i \leq 1$
if $\bar{b}_i, \bar{w}_i$ are in the interval $0\ldots 1$.
Furthermore, the case $\bar{b}_i=\bar{w}_i=0$ is not possible. \vspace*{3pt}

\noindent\parbox{0.22\textwidth}{\ \ The following position in 
diagram \ref{dia3b} has three fixed points of which
the meaningful one is stable.  We assume that the surrounding blocks are
alive, i.e.\ $s_{c3}=s_{d3}=1$.  The probability \vspace{3pt} of
}
\hspace{0.01\textwidth}
\parbox{0.24\textwidth}{
\[ \myblock {0.15\textwidth} {
  \board1--7/1--4/
  \goboardini\alphatrue
  \whites{C1C2D3E3F3F2F1}
  \blacks{B1B2B3C3D2E2E1}
  \purenumbers[1]{D1}
  \shipoutgoboard
 }{}{}{dia3b}\]
} 
the point $d1$ to have only black neighbours is 
$s_{d2}(1-s_{c1})$ and thus the probability $\bar{w}_{d1}$ to have
at least one white neighbour is $\bar{w}_{d1}=1-s_{d2}(1-s_{c1})$.
Similarly, we get $\bar{b}_{d1}=1-(1-s_{d2})s_{c1}$ which
inserted into equation (\ref{e3}) gives
\[w_{d1}=\frac{s_{c2}s_{d2} - s_{d2} + 1}
              {2s_{c2}s_{d2} - s_{c2} - s_{d2} + 2} . \]
The probability for the block at $d2$ to be captured is
$s_{c1}w_{d1}$. Thus 
\[ s_{d2} = 1 - s_{c1}w_{d1}  \]
and similarly
\[ s_{c1} = 1 - s_{d2}b_{d1} .  \]
Together with $w_{b1}+b_{d1}=1$ this is the system to be solved.
It has the 3 solutions
\[b_{d1}=w_{d1}=\frac{1}{2}, \ \ s_{c1}=s_{d2}=\frac{2}{3} \ ;\]
\[b_{d1}=0, \ \ w_{d1}=1, \ \ s_{c1}=1 \ \ s_{d2}=0 \ ;\]
\[b_{d1}=1, \ \ w_{d1}=0, \ \ s_{c1}=0 \ \ s_{d2}=1 \]
of which the first is stable and the others are the
extreme solutions mentioned above. What seems to be inaccurate 
in the first solution is $s_{c1}=s_{d2}=\frac{2}{3}$ instead of
$\frac{1}{2}$.

In the computation of $\bar{w}_i, \bar{b}_i, s_j$ we made 
the following systematic error.
If two events $A, B$ have probabilities $p_A, p_B$ then the
probability that $A$ {\em and} $B$ occur is equal $p_A p_B$ 
only if $A$ and $B$ are independent.
In the above computation we said that the probability for the 
block at $d2$ in diagram \ref{dia3b} to be captured is
$s_{c1}w_{d1}$. But if White plays on $d1$ then the strength of the
white block on $c1$ increases and is not anymore $s_{c1}$, so both
probabilities $s_{c1}, w_{d1}$ are not independent.

A more clear cut example is shown in
diagram \ref{dia3c} where the external blocks with stones on $d5, e4$ 
are assumed to be alive.\vspace*{-3pt}

\noindent\parbox{0.26\textwidth}{The probabilities $w_{c1}, w_{c2}$ for White to
move on any one of the two points $c1, c2$ is initialized to 0.5 as for
all moves and it stays $>0$ during the iterations. But the real
probability for White to go on both points in a game is zero due to
the suicide rule in Go.}
% 6 | + + + + + 
% 5 | O O O + +
% 4 O O # # # + 
% 3 O # O O # + 
% 2 # # + O # + 
% 1 +-#-+-O-#-+-
%   A B C D E F 
\hspace{0.01\textwidth}
\parbox{0.2\textwidth}{
\[ \myblock {0.15\textwidth} {
  \board1--6/1--6/
  \goboardini\alphatrue
  \whites{A3A4B4B5C5D5D1D2D3C3}
  \blacks{A2B1B2B3C4D4E1E2E3E4}
  \shipoutgoboard
 }{}{}{dia3c}\]\vspace*{3pt}
} 

Although the systematic error of ignoring correlations between moves
seems grave, the resulting numerical error is not. SEDS obtains
$s_{b1}=0.98, s_{d1}=0.53$ which (given the scale of values obtained
by SEDS in races of liberties) is a clear indication that the black block
is alive and the white block is dead. This is a good example showing
that SEDS is a robust method providing suitable indicators and
data for more specialized semeai and life \& death investigations.

A good question is how many iterations are necessary for all values to
settle down so that all changes are less than some threshold
parameter, say $10^{-5}$.  The interesting result is that for clear
cut situations only few iterations $(<10)$ are necessary whereas for
very unstable situations (for example, two attached blocks both under
atari) the number of iterations can reach hundreds or thousands. This
opens the possibility to get as a by-product a measure of instability.
On the other hand determining unstable regions this way requires more
iterations than would be needed for only a strength estimate.

In defence of the approach it should be mentioned that when
establishing a ranking of moves often not the absolute strength values
matter but the relative values.

\subsection{Statistics on Professional Games} \label{stat}

The SEDS function has been tested in detail by trying to predict the
next move in professional games. Alternatively one can also look at
these tests as tries to exclude as many moves as possible apart from the
professional move if the main intention is to use SEDS to narrow
the search of MCTS.

The test consisted of doing a 1-ply search for all positions occurring
in all 50,000 professional games from the GoGoD collection
\cite{GoGoD2007}. For each board position in these games this includes 
\begin{itemize}
\item performing each legal move in this position, 
\item iterating the dynamical system in the new position until the
  system stabilizes,
\item adding up all probabilities of blocks to survive and points to
  be owned by either side and thus reaching a total score,
\item ordering all legal moves according to their total scores,
\item finding and recording the position of the professional move in
  the ranking of all moves.
\end{itemize}
The statistics have been recorded separately for each move number
because at different stages of the game the static evaluation has
different strengths and weaknesses.

\begin{figure}[ht]
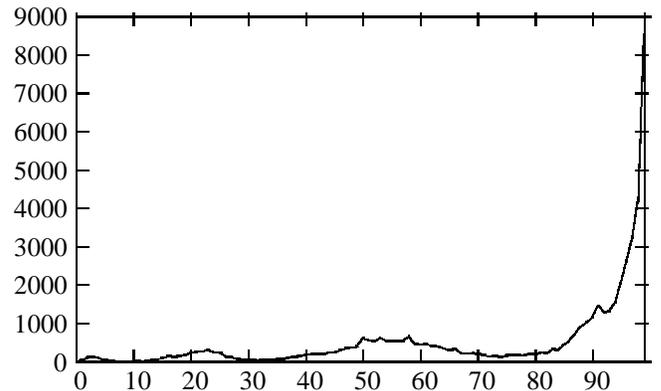
 
\include{stat9b-50}
\vspace{-18pt}
\caption{A statistics of the ranking of the next professional move
according to SEDS in all positions with move number 50 from 50000 
professional games} \label{fig1}
\vspace{-6pt}
\end{figure}

The results of this test are reported under {\tt \small
http://lie. math.brocku.ca/twolf/papers/WoSE2010/2} due to the size of
the diagram files. The data have been produced by evaluating 5.4
million positions from the 50,000 professional games of the GoGoD
collection (\cite{GoGoD2007}).  On this web site three sequences of
diagrams are shown. Each sequence contains over 400 diagrams, one for
each move number. Figure \ref{fig1} is one of the diagrams of the
first sequence. It shows the number of positions in which the
professional moves made in those positions land 
in the range $x\ldots (x+1)\%$ of legal moves
as sorted by the SE where $x=99$ are the top moves falling in the
$99\ldots 100\%$ range and $x=0$ are the worst moves.  Thus the higher
the graph is on the right and the lower it is on the left, the better
is the static evaluation.

\begin{figure}[ht]
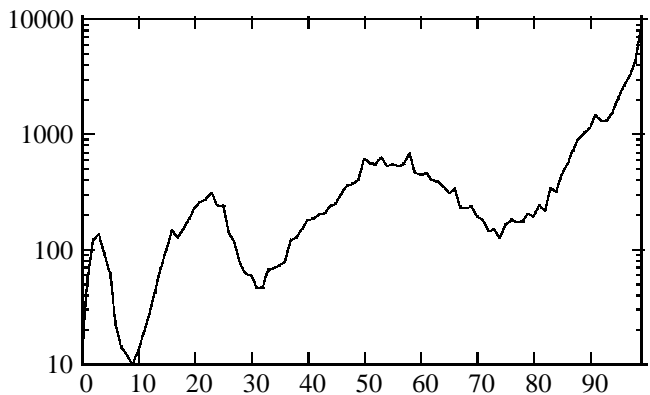
 
\include{stat9blog-50}
\vspace{-18pt}
\caption{The diagram of figure \ref{fig1} here with logarithmic vertical
axis} \label{fig2}
\vspace{-9pt}
\end{figure}

The second series of diagrams differs from the first by having
a logarithmic vertical axis. This is useful if the emphasis is to
safely ignore moves from further consideration in MCTS because then we
want to be reasonably sure that SE does not rank good moves (moves
played by the professional player) as bad (on the left side of figure
\ref{fig1}). In other words, we want to be sure that the graph is low 
on the left and to highlight that range a logarithmic vertical scale is 
useful. Figure \ref{fig2} is the logarithmic version of figure \ref{fig1}).

\begin{figure}[ht]
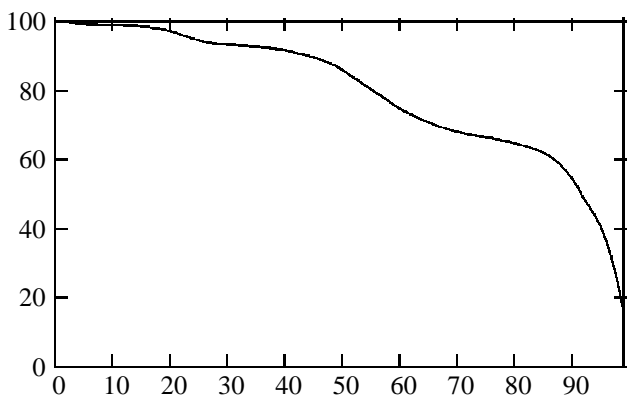
 
\include{stat9bacc-50}
\vspace{-18pt}
\caption{The data of figure \ref{fig1} here in a cumulative form of a
survival function} \label{fig3}
\vspace{-9pt}
\end{figure}

If one normalizes the vertical axis in figure \ref{fig1} then this curve is
the probability density $P(x)$ of the ranking of the professional move 
among all moves. If one accumulates this density from the right one obtains
a so called 'survival function' $R(x)$: % $R(x):=\int_x^{99} P(u) du$
$R(x):=\sum_{u=x}^{99} P(u)$ which is displayed in figure \ref{fig3}.
For example, a point on the graph with horizontal coordinate 85
and vertical coordinate 62 means: The professional move is kept with
a probability of $62\%$ (it survives) if the worst $85\%$ of the moves are
dropped (worst according to the SEDS).

\subsection{Interpretation}
In view of the simplicity of the static evaluation function SEDS the
results as shown in figure \ref{fig1} are surprisingly good.

Deficiencies are not difficult to explain. Because of the local
concept of the dynamical system approach the SEDS has no concept of
life (except a hard-wired fast recognition of static life), i.e.\ it
does not know of the need of two eyes and the benefit of destroying
eyes.  For the SEDS in its current form (Dec 2009), strength is 100\%
correlated with resistance against being captured. As a consequence, 
sacrifice moves, like Black on A in diagram \ref{fig4}, get a low
ranking. This is an extreme example where the professional move (Black
on A) gets the lowest ranking of all moves by SEDS. Moves of this type
make up the leftmost hump in figure \ref{fig2}.

Whereas the humps on the left of figure \ref{fig2} are due to good
(professional) moves getting a low evaluation by SEDS, the dents on
the right of the graph are due to bad moves getting a high evaluation
by SEDS.  The rightmost dent in figure \ref{fig2} is due to the
feature of the evaluation function to favour moves on the $2^{\rm nd}$
line, especially the 2-2 points (e.g.\ on a $19\times 19$ board the 
points b2,b18,s2,s18). Again, this is a consequence of not
knowing about the need for 2 eyes due to not knowing that Black and
White can not do 2 moves at once, i.e.\ fill 2 
\vspace{-6pt}
%\begin{figure}[ht] 
\[
\myblock {0.45\textwidth} {   % <-- new
\board1--19/1--19/
\goboardini
\alphafalse
\whites{D4D17P2P3Q1R2R3Q4Q5}
\blacks{D15P4P5Q6Q16R4S5}
\letters{Q3}
\shipoutgoboard 
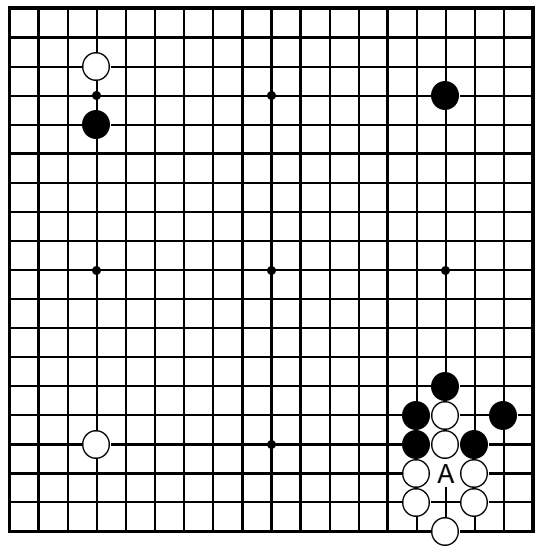
}{}{A position from a professional game where SEDS fails       % <-- new
    due to lacking a concept of eyes.}{fig4}                   % <-- new
\]
%\vspace{-18pt}
%\caption{A position from a professional game where SEDS fails 
%         due to lacking a concept of eyes.} 
%\label{fig4}
%\vspace{-3pt}
%\end{figure}

\noindent eyes at once. For the
SEDS a move on a $3^{\rm rd}$ or $4^{\rm th}$ line can be cut under by
the opponent on the $2^{\rm nd}$ line because SEDS does not know that
the move on the $2^{\rm nd}$ line needs two eyes.

When skipping through diagrams on {\tt \small http://lie.math.
brocku.ca/twolf/papers/WoSE2010/2} one sees that the current version
of SEDS is most useful early in the game when life \& death fights do
not play a big role yet but also after about 15 moves when the exact
influence of the edge of the board is not so crucial anymore.

Based on these findings it is expected that an appropriate
consideration of life based on 2 eyes when computing the strength of
blocks will lead to a significant improvement of SEDS. The problem is
to find a natural merger of the need for 2 eyes with the current
interaction formula (\ref{e3}). Also, the concept of life is not local,
so the solution of the dynamical system and the determination of a
(non-local) measure of life based on 2 eyes must be merged naturally 
into one algorithm. Of course, one could make quick progress with a
superficial repair but the aim of this exercise is to get a lasting
concept that has no artificial parameters and no artificial constructs
and thus scales (can improve indefinitely with increasingly available
computer power) and thus has the potential to result in a strong
program in the long run.

\subsection{Timing}
The following times have been recorded on a Dell Optiplex GX620 PC
with Intel(R) Pentium(R) D CPU 3.40GHz processor with 2MB cache size
running Linux. One CPU was used.  Times reported in table 1 are the
result of averaging the times for evaluating 
% 400 positions from 400 professional games from the GoGoD game collection 
% \cite{GoGoD2007} each position on the same move number. 
400 positions for each move number given in the table, each position 
from a different professional game from the GoGoD game collection 
\cite{GoGoD2007}.

The computations in column 2 include making a move and updating the strength of
blocks and influence at points on the whole board. In column 3 these steps are 
made for each legal move.
\begin{center} 
\begin{tabular}{|c|c|c|} \hline 
move number & time in $\mu$s for & time in ms for \\
            & static evaluation  & ranking moves  \\ \hline 
   10       &       46.2         &     16.2       \\ \hline 
   30       &       51.8         &     17.1       \\ \hline 
  100       &       73.8         &     19.2       \\ \hline 
  130       &       83.9         &     19.3       \\ \hline 
  200       &      104.4         &     16.7       \\ \hline 
  300       &      180.0         &     10.8       \\ \hline 
\end{tabular} \vspace{6pt}\\   
Table 1. Average times for static evaluation and ranking of moves.
\end{center}               
As the update of the strength of a block is slower to compute than an
update of the influence at a point, the static evaluation (column 2)
becomes slower as more stones are placed on the board. On the other hand,
the more stones there are on the board, the fewer legal moves exist and
ranking all moves (column 3) becomes faster.

\subsection{Parameter Dependence}
The current form of SEDS has two parameters.
One parameter {\tt stop\_value} is the minimal change of
influence and strength which keeps the iteration alive, i.e. if the
influence in a point or the strength of
a block changes by at least this amount then the neighbouring points and
blocks have to be iterated again.
The other parameter {\tt max\_iter} is the maximum number of
iterations for any point or block.
When testing {\tt stop\_value} $=10^{-2\ldots -7}$%, 10^{-3}, 10^{-5}, 10^{-7}$
and {\tt max\_iter} values $=2\ldots 10,100,1000$ the following was found.
\begin{itemize}
\item 
The {\tt max\_iter}
value has hardly any effect, neither on time, nor on accuracy
because the number of iterations is typically very low ($\leq 5$). Rare
exceptions occur in beginner games when several blocks with only one 
liberty are neighbour to each other and thus the situation is
extremely unstable and many iterations between such blocks would occur. 
\item A {\tt stop\_value} of $10^{-3}$ gives the best performance/cost ratio
in predicting moves in professional games. 
Lowering this value to $10^{-5}$ adds another 25\% to the computing time 
and lowering it to $10^{-7}$ adds again 15\% without improving the 
prediction of professional moves noticeably. Increasing the {\tt stop\_value} to 
$10^{-2}$ gives only very little time savings but lowers performance.
\item An important conclusion is that allowing at least 3 or 4 iterations per item 
(point or block) does improve predictions. This is an indirect proof that SEDS 
recognizes, for example, when a white block is strong because a 
neighbouring black block is weak because a neighbouring white block is 
strong. 
\end{itemize}

To summarize, {\tt max\_iter} is a parameter in the program but 
not a parameter for the computation 
that strikes a balance between accuracy and complexity.
{\tt stop\_value} has an effect on accuracy and complexity but only a minor one.
It is a 'natural' parameter in contrast to what one may call 
'artificial' parameters, like
\begin{itemize}
\item the size of fixed size pattern that are used,
\item ad hoc parameters that classify the strength of blocks like a threshold 
      number of liberties such that a block is considered alive, 
\item a fixed size distance of the last move which favours follow up
      moves in its neighbourhood.
\end{itemize}
The dependence of accuracy and complexity of computation on such
parameters is very uneven. If one would vary such a parameter and plot
the resulting accuracy versus the computational complexity of the
calculation then the curve would be flat and rise very slowly for low
and medium complexity and would only for very large computational
complexity rise more steeply and reach higher accuracy.

In contrast, 'natural' parameters have a more direct relation to the accuracy
of the computation, like 
\begin{itemize}
\item parameters regulating a genetic learning algorithm, or the speed of lowering 
      the temperature of simulated annealing,
\item the number of terms of an Fourier or Taylor series expansion of a function,
\item the number of simulations of MCTS,
\item the {\tt stop\_value} parameter for iterations of SEDS.
\end{itemize}

\section{Comparisons with other static evaluations} \label{comparisons}
\subsection{Criteria for Static Evaluations}
Because there is ambiguity of what one can consider as an evaluation
function, there are many criteria that matter for their characterization.
%It is hardly possible to compare different static evaluation functions
%especially if they are based on different principles.
An evaluation function is good if it
\begin{enumerate}
%1
\item ranks top and good moves high,
%2
\item ranks bad moves low,
%3
\item is fast, 
%4
\item is flexible in time management, i.e.\ it is able to find reasonable
      quality moves if only little time is available and better moves if
      more time is available,
%5
\item knows about the value of the next move in order to take more time 
      and be more accurate if much is on stake, 
%6
\item knows when its own predictions are less accurate (e.g.\ in an unstable
      region where search is the proper tool) or more accurate (in a stable 
      region),
%7
\item knows about the risk involved in a move to select safer moves when 
      being ahead and more risky moves when being behind,
%8
\item knows about the strengths of the search program and thus gives moves 
      a high priority if they lead to positions which the search can handle 
      well,
%9
\item performs equally well for positions resulting from skillful play
      (professional games), resulting from beginner play or from high 
      handicap games,
%10
\item provides, as a side product, data about points and blocks on the board,  
      date which are re-usable in the computation of higher concepts,
%11
\item has potential for improvement, e.g.\ if extensions can be added 
      efficiently, for example, the recognition of safe links or of more 
      general semeai.
\end{enumerate}

\subsection{Comparisons}
\begin{figure}[ht]
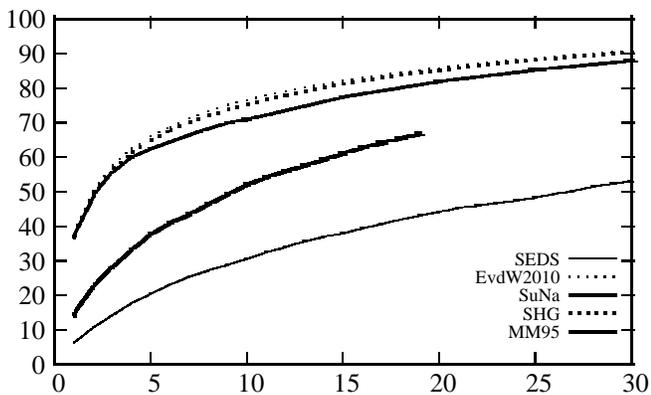
 
\include{stat9e-rnk}
\vspace{-18pt}
\caption{A statistics of the ranking of the next professional move
by different evaluation functions}
\label{fig0}
\vspace{-6pt}
\end{figure}

The first ability is easily tested by predicting moves of professional games.
Figure \ref{fig0} compares SEDS with other programs.

The curves EvdW10 (\cite{EvdW10}), SuNa (\cite{SuNa08}), SHG
(\cite{SHG06},\cite{Stern08}) are close. They are produced by programs
that match pattern learned from professional games but newer versions
match also other features (see \cite{Stern08}). Such programs are the
best in predicting professional moves, especially in the opening. They
are also reasonably fast with typically 0.3-100 ms for a full board
ranking. But because of weaknesses in criteria 4-10 and partially 11
they are not suitable for playing complete games. Instead, pattern
matching programs are useful to provide prior knowledge to MCTS
especially in the opening, as applied, for example, in the program
Steenvreter from Erik van der Werf (\cite{EvdW10}).

For criterion 2 no tests are currently available. It would be interesting to
see how pattern matching programs perform in positions resulting from
beginners play which may show only small patterns learned from professional
games and which consequences this has on the evaluation of bad moves.

It may be unexpected that SEDS, which on a first look seems to be
extremely slow, is in fact comparable to or even faster in speed than
SuNa (\cite{SuNa08}) and of same speed as the pattern matching
programs described in \cite{WUPH03}, \cite{Stern08}, although slower
than EvdW10 (\cite{EvdW10}) (according to personal communication with
the authors). In section 3.4\ of \cite{CWHUB08} it is commented that
the time for computing the heuristic value of a move in the program
{\sc Mango} (mainly consumed by pattern matching) is in the order of 1
ms which would be about 10 times slower than the times shown in column
2 of Table 1.

In predicting professional moves SEDS is clearly weaker than the pattern
matching programs but its strength is that it has good potential for criteria
6) (the local number of iterations turns out to be a good indicator for the
stability of a block or region), 
7) (because SEDS has a numerical value for a score and a measure of
instability),
9) (SEDS is not based on pattern learned from a specific class of positions),
10) (SEDS generates and stores all neighbourhood relations because it needs
them frequently and it provides influence values at points and strength 
values of blocks), and
11) (higher concepts are easier to be built on top of rich
available basic data).

The graph MM95 in figure \ref{fig0} is the cumulative version of the
figure in section 8.2 of Martin M\"{u}ller's PhD thesis \cite{MMPhD95}
produced by the program Explorer. It is an elaborate knowledge based
program \cite{MM02}, thus it naturally scores better in move
prediction than SEDS which is based purely on neighbourhood
relations. It would be interesting to compare the time consumption of
the Explorer evaluation function and SEDS.

Another interesting study would be to simulate an evaluation function
by a MCTS based program that is given only 1-10 ms time. MCTS has
advantages in criteria 4, 7, 8, 9 and partially 10, 11 compared to
pattern matching programs.

%In its current form SEDS does not seem competitive with pattern matching
%in predicting  professional moves in professional games 
%but it so far just a 
%has a number of other advantages.
% much potential because it is only...

%\section{Summary and Future Tasks} \label{future}
\subsection{Characterization of SEDS}
The dynamical system as formulated in section \ref{dynsys} is a first
version that allows us to study general properties of such an
approach.  This version is minimal and has no concept of 100\% safe
liberties (eyes) and thus has no concept of life in
general\footnote{As a start, static life which never has to answer any
move has been implemented.} only a concept of strength as the
resistance against being captured and a concept of influence at empty
points.  As move prediction experiments show, these concepts already
capture a good part of the nature of Go.

The value of SEDS lies in providing a new source of Go knowledge that 
can be experimented with and potentially used to create prior knowledge
for MCTS programs. The following are the strengths of SEDS in its current 
form.
\begin{itemize}
\item The provided knowledge is of global nature because all 
  points and blocks are coupled to their neighbours and so on. For example,
  one block 'feels' the weakness of an opponent neighbouring block due to the
  strength of an own block behind that.
\item SEDS is applicable to any board position and it is robust in the 
  sense that approximations in the
  assumptions, like the non-correlation of future moves, do only disturb
  numerical results, and do not make them completely wrong.
\item In all tested positions iterations lead to a single
  fixed point of the dynamical system which describes the strengths of 
  all blocks and the influence at all points. The fixed point reflects the 
  situation on the whole board, no artificial local cutoff is assumed 
  which is, for example, the case when local pattern are used.
\item The computation of fixed points is reasonably fast because 
  only few iterations are needed and an incremental form of the
  computation is possible which limits iterations to areas where
  a move that has been made has an effect on its neighbourhood.
  Consequently, the speed of SEDS is comparable to that of pattern 
  matching programs and thus probably faster than traditional 
  knowledge based programs. The ranking of moves based on the SEDS scores 
  they reach, can be parallelized and a more heavy SEDS (e.g.\ by including 
  life \& death computations) makes the parallelization more coarse grain 
  and more effective.
\item The influence radiated from blocks depends on their strength and the
  block's strength depends on the influence at neighbouring points.
  None is artificially fixed. 
\item The special situation of points and blocks at the boundary 
  is taken care of automatically due the naturally reduced number of 
  neighbouring points and blocks.
\item The SEDS described in this paper is pure in the sense that it is free
  of artificial parameters and thus has potential to be extended and tuned.
\item Individual components of the SEDS can be modified independently,
  like the dependence of the strength of a block on the influence of
  neighbouring points or on the strength of neighbouring blocks or
  like the dependence of influence at a point on the strength of
  neighbouring blocks or influence at neighbouring points.
\item As a by-product of SEDS information on the strength of blocks and the
  influence at points is obtained that can be useful for other specialized
  modules, like tactical search or generalized strength pattern.
\item Weaknesses of SEDS can be determined easily by doing move prediction
  experiments and filtering out positions where the professional move ranks
  low in the SEDS ranking.
\end{itemize}

\section{Future Tasks} \label{future}
Three directions of future work are: to improve the computation of the strength
of blocks and the propagation of influence, to make good use of
the computed values and to merge static evaluation with MCTS. 

\subsection{Improvements of the Algorithms} \label{improvements}
The following are possibilities to improve this first version of SEDS.
\begin{itemize}
\item The formula for the strength of a block could be improved by 
  giving the number of liberties a weight in addition to the 
  currently used influence values at the liberties.
\item An inspection of worst performances in move predictions showed
  that the concept of life, i.e.\ the need for two eyes is not automatically
  covered by this first version of SEDS and needs to be
  added. A possibility would be to perform iterations in two stages.
  After a first sequence of iterations settled down a quick life and death
  analysis using available influence values could be performed and
  resulting strength values of blocks be considered in a second series
  of iterations.
\item Another non-static concept based on moves are safe links. These are
  also not automatically covered by SEDS and have to be incorporated in 
  some organic way (see section \ref{multipole} below for a suggestion).

  The reason for not including eyes and links already into this initial 
  version of SEDS was to study at first a pure version of dynamical system
  before extending it. The purpose of minimizing the number of concepts
  and parameters is to keep the program scalable in the long run.
\item SEDS should provide a local awareness of stability (i.e.\ the
  dependence of the local strength measures on who moves next), of the
  size of an unstable area and thus of the importance of investigating
  moves in this area. If this measure proves to be a good heuristic
  for the urgency to investigate moves through MCTS then the 1-ply search 
  to rank moves would not be needed anymore.
  This would result in a significant speedup.
\item Because pattern matching and SEDS are complimentary in their
  approach, a combination of both has good chances to be better than
  each one of them. 

  In pattern matching the shape of blocks matters whereas the strength
  of blocks touching the pattern edge and potentially extending beyond
  the edge is unknown.  Differently in SEDS, shapes of blocks are
  ignored but approximate strength values are known. Pattern are local
  objects whereas strengths and influences computed by SEDS are global
  objects.

  In a first approach of merging pattern matching and SEDS one could
  simply add both move recommendations with appropriate weights. The
  next stage would consider pattern enriched with strength and
  influence values. Such pattern could be smaller and still have high
  predictive value.

\end{itemize}

\subsection{Multipole Moments of Influence} \label{multipole} 
In this section ideas are given how to utilize an influence field. 
 
The philosophy behind SEDS is to model the 'real strength' of blocks and 'real
influence' as close as possible to what strong human players think about
that. If the resulting influence field would be a proper model to describe
what is going on on a Go board (i.e. if it would have some intrinsic meaning
for Go) then to some extend also the components of a multipole expansion of
that field (i.e. the discrete analogue of that) should have a meaning in
Go. Multipole moments describe changes of a field in the neighbourhood of a
point.
 
The first term of such an expansion (monopole component) is the value of
the field itself in a point. It is an indicator of the likelihood of a point 
to be owned by either side at the end of the game. 
 
The second term (dipole moment, for continuous fields known as gradient)
indicates the direction of fastest increase of influence and the rate of
change of influence in this direction and could be used as indicator how to
reach or avoid a strong block of a specific colour, or how to run out in the
open to gain liberties.

\noindent\parbox{0.35\textwidth}{
\ \ The third term (quadrupole moment) contains second order differences. What
is of interest from these is a measure of how much the point is a
saddle point and can be used as \ an indicator how important the point \ is} 
\hspace*{0.01\textwidth}
\parbox{0.11\textwidth}{
\[ \myblock {0.15\textwidth} {
  %  + a +
  %  b x c
  %  + d +
  \board5--7/5--7/
  \goboardini\alphafalse
 \letters{f7e6g6f5}
 \letters[24]{f6}
  \shipoutgoboard
 }{$\!\!\!\!\!\!\!\!\!\!$}{}{dia3d}\]\vspace*{3pt}
}
for connecting or cutting blocks.
For example, in Diagram \ref{dia3d} let $A,B,C,D$ be influence values at these points
(or strength values of blocks with a stone at these points),
all being neighbours to the point $X$  
then $|A+D-B-C|$ is a good indicator for the urgency to connect own blocks and/or
separate and cut opponent blocks. The formula could be refined by including
diagonal neighbours.

% \subsection{Block Strength Identification}
% As mentioned, for example, in \cite{KC01a} there are the following aspects of
% strength for a block: semeai, territory, connection and life-and-death.
% \begin{itemize}
% \item Preliminary tests show that SEDS is very useful and rather accurate for a
% static evaluation of semeai because for semeai the approachability of
% liberties matters which is determined through the iterations in SEDS.  
% \item The
% influence values computed in SEDS are also a good basis for estimating
% territory and thus strength based on territory. 
% \item Ideas of how to use influence
% for making moves that establish or cut connections are given in section
% \ref{multipole}. 
% \item Compared to semeai, a life-and-death fight has the
% additional component of sacrificing blocks to prevent splitting eyes and to
% kill (nakade moves). SEDS is not well suited to anticipate such moves
% statically. One reason is that sacrificed blocks get a low strength value in
% SEDS because they can be captured and thus the surrounding chain gets a higher
% strength value. The other reason is that shape and size of blocks do matter if
% they are to be sacrificed but both are not recorded in SEDS. A tool to
% compensate this weakness can be a database of monolithic eyes \cite{mono}.
% Another measure is the 2 stage computation of block strength as outlined
% in section \ref{improvements}.
% \end{itemize}

\subsection{Merging Static Evaluation and Treesearch}
Whereas MCTS is self-sufficient for game playing, evaluation functions are
not. When using SEDS to make moves directly to play games 
as program {\tt moag-0.3} on CGOS (\cite{CGOS}) 
% in March 2009 ELO rating 600-700
all games were lost with the exception of two wins against {\tt AmiGoGtp}.

The merger of MCTS and an improved SEDS will be a main future task and
a research project on its own. In doing this one would want to be able
to vary smoothly the times allocated to both, at first statically then
dynamically depending on the board position.

The computational cost of SEDS is more justified in the opening when
precise MCTS is very expensive and in non-fighting positions when SEDS
is more accurate. Investing time in SEDS will be less justified
towards the end of the game when MCTS does a perfect job and in
all-or-nothing fights when SEDS is not accurate enough due the
occurrence of sacrifice moves where the shape of throw-in blocks
matters which is disregarded in SEDS. As a first attempt one could
add recommendations for moves from MCTS and SEDS as described in 
formula (3.2) in \cite{CWHUB08}.

A run-time library under Linux is available that provides influence values at
points and strength values of blocks. It also has a function which ranks all
legal moves by their estimated quality. To try it out within a MCTS program
please contact the author.

\section{Summary} \label{summary}
Based on the local nature of the capture rule in Go, which is
currently not taken advantage of in MCTS, it is argued that a
dynamical system (DS) is suitable to describe a board position in Go.
It is shown how minimal natural assumptions lead to the formulation of a DS 
that has a number of useful properties.

After a discussion of principal limitations of a local dynamical
system approach the results of move prediction tests in professional
games are described.Given the simplicity of the currently
investigated SEDS not using any Go-knowledge from high level games, no
pre-computed information, no tactical information or algorithms other
than static life and death its move prediction abilities are
surprisingly good. In comparisons with other static evaluations
strengths and weaknesses of SEDS are described.

The characteristics of SEDS and the many possible improvements listed
in the paper give the new approach much potential that should be
explored.

\section*{Appendix}
In this appendix an example is given to support the claim that in Go
there is a field that embodies strength and influence, which is
not an artificial human construct but which is at the heart of
the game and is of intrinsic, fundamental nature. In the following
position such a hypothetical field is used to explain the optimal move
which is a sacrifice. 
\[\goboardini
\alphatrue
\parbox{0.21\textwidth}{
 \myblock {0.183\textwidth}{
  \board1--8/1--10/
  \whitetris{B5}
  \whitecrosses{B2}
  \whites{D2D3C2D4E3D5B5C6C7C9D8}
  \blacksquares{B3}
  \blacks{A2C3C4C5D6D7E2E4E5E6F2F3G4}
  \shipoutgoboard
 }{}{\\ \hspace*{6pt} \tbpstone\ to move.}{dia5}
} \ \ \ \ \ \ 
\parbox{0.21\textwidth}{
 \myblock {0.183\textwidth}{
  \blackfirsttrue
  \board1--8/1--10/
  \whites{D2D3C2D4E3B2D5B5C6C7C9D8}
  \blacks{A2B3C3C4C5D6D7E2E4E5E6F2F3G4}
  \numbers{B6B4A3B7B1}
  \shipoutgoboard
 }{}{\\ \hspace*{6pt} \tbpstone\ lives.}{dia6}
}
\]
In diagram \ref{dia5} the aim of Black is to make its corner block
\tbsstone\ alive which can only happen by capturing one of the white
blocks \twtstone\ , \twcrstone\ . But neither one can be captured
directly by playing only in its vicinity (\tbpstone\ on b6 would be
captured by \twpstone\ on b7 and \tbpstone\ on the first row would be
answered by \twpstone\ on b4 and would be too slow). Nevertheless, a static
evaluation function modeling an influence field, for example the one 
described in section \ref{dynsys}, would give small but
nonzero values for Black's strength around a6, b6 and on the first row
at b1, c1, d1. If modelled correctly these influence values alone
should be too low to indicate the death of one of the white blocks
\twcrstone, \twtstone\ individually but if all influence is added up
and increased by a value equivalent to the advantage of having the turn
then the total should be enough to indicate life for \tbsstone\ .

The remaining question is how to convert this prospect of life for Black
into a good first move. The influence field also helps here.

If this field is to be meaningful then it should not change erratically
from one move to the next, except at the very end when the position
becomes completely settled and the value jumps to one of the two extremes.
So Black can not expect to change the total sum of all Black strengths 
through a clever move. But what Black can expect to do is to shift the
distribution of its influence. To succeed in this example, Black needs to
bundle all its influence onto its weakest white neighbour block, which is 
\twcrstone\ , i.e.\ to move its small influence at a6, b6 towards 
the block \twcrstone\ to weaken it further. The result of the sequence in
diagram \ref{dia6} is that White now has total control of the points
a6, b6 and Black in exchange gets one extra move towards capturing the block 
\twcrstone\ which is enough in this case.

This principle of 
%- identifying own minor influence which could be utilized, i.e.\ 
%
%\item finding a good sacrifice move which 
%
%\item poses a real thread, i.e.\ if not answered it can capture or cut,
%\item takes one liberty of an opponent block with only few liberties,
%      and/or performs a cut on this block,
%\item is not captured in one move,
bundling influence to overcome a threshold strength in a local target
area in order to live, link, kill or cut, explains all sacrifice
moves, not only in this example and not only in life and death
problems. This is a good guidance not only for players but also for
computer programs to formulate tactical and strategic aims, on one
hand being bound by a stable total sum of influence but on the other
hand being allowed to shift influence and to focus it to overcome
threshold values locally for living and killing. These threshold
values result from the discrete requirement of having two eyes for life and
the discrete nature of the capture rule of capturing all stones of a
block at once.

A variation of this principle is to look for moves which make own weak
stones good sacrifice stones, aiming to move the remaining strength of these
stones to a distant and more important area.

In some games the principle of collecting thinly spread potential through
playing 'light', i.e.\ through playing stones which have a good chance of
being sacrificed later, is not only a tactical concept but a strategic
one. When a professional player gives a strong amateur player many handicap
stones then Black can not simply be fooled, the only way for White to win is
to collect all potential on the board by playing light and probably 
sacrificing stones.

But also in even games thinly spread influence/potential is an
important concept. Influence may be shifted around to force the
opponent to become very strong on one side of a local area in order to
gather own strength and be better prepared to attack on the opposite
side of that area, in other words to create imbalance in the opponents
position. This is known as a proverb: Attach against the stronger
stone (for example, in\cite{OH92}, p.\ 121).

\section*{Acknowledgements}
Work on this project was supported by a DARPA seedlings grant. The
author thanks the other members of the funded group from the Stanford
Research Institute for discussions, especially Sam Owre for binding
SEDS into the Fuego program and running tests on CGOS. Bill Spight is
thanked for comments and discussions. Computer tests were run on a
Sharcnet cluster {\tt http://www.sharcnet.ca}.

\newpage
\bibliographystyle{IEEEtran} \bibliography{public}

\end{document}

%% file: stat9b-50.tex
% GNUPLOT: LaTeX picture
%\setlength{\unitlength}{0.240900pt}
\setlength{\unitlength}{0.168pt}
\ifx\plotpoint\undefined\newsavebox{\plotpoint}\fi
\sbox{\plotpoint}{\rule[-0.200pt]{0.400pt}{0.400pt}}%
\begin{picture}(1500,900)(0,0)
\sbox{\plotpoint}{\rule[-0.200pt]{0.400pt}{0.400pt}}%
\put(160.0,82.0){\rule[-0.200pt]{4.818pt}{0.400pt}}
\put(140,82){\makebox(0,0)[r]{ 0}}
\put(1419.0,82.0){\rule[-0.200pt]{4.818pt}{0.400pt}}
\put(160.0,168.0){\rule[-0.200pt]{4.818pt}{0.400pt}}
\put(140,168){\makebox(0,0)[r]{ 1000}}
\put(1419.0,168.0){\rule[-0.200pt]{4.818pt}{0.400pt}}
\put(160.0,255.0){\rule[-0.200pt]{4.818pt}{0.400pt}}
\put(140,255){\makebox(0,0)[r]{ 2000}}
\put(1419.0,255.0){\rule[-0.200pt]{4.818pt}{0.400pt}}
\put(160.0,341.0){\rule[-0.200pt]{4.818pt}{0.400pt}}
\put(140,341){\makebox(0,0)[r]{ 3000}}
\put(1419.0,341.0){\rule[-0.200pt]{4.818pt}{0.400pt}}
\put(160.0,428.0){\rule[-0.200pt]{4.818pt}{0.400pt}}
\put(140,428){\makebox(0,0)[r]{ 4000}}
\put(1419.0,428.0){\rule[-0.200pt]{4.818pt}{0.400pt}}
\put(160.0,514.0){\rule[-0.200pt]{4.818pt}{0.400pt}}
\put(140,514){\makebox(0,0)[r]{ 5000}}
\put(1419.0,514.0){\rule[-0.200pt]{4.818pt}{0.400pt}}
\put(160.0,601.0){\rule[-0.200pt]{4.818pt}{0.400pt}}
\put(140,601){\makebox(0,0)[r]{ 6000}}
\put(1419.0,601.0){\rule[-0.200pt]{4.818pt}{0.400pt}}
\put(160.0,687.0){\rule[-0.200pt]{4.818pt}{0.400pt}}
\put(140,687){\makebox(0,0)[r]{ 7000}}
\put(1419.0,687.0){\rule[-0.200pt]{4.818pt}{0.400pt}}
\put(160.0,774.0){\rule[-0.200pt]{4.818pt}{0.400pt}}
\put(140,774){\makebox(0,0)[r]{ 8000}}
\put(1419.0,774.0){\rule[-0.200pt]{4.818pt}{0.400pt}}
\put(160.0,860.0){\rule[-0.200pt]{4.818pt}{0.400pt}}
\put(140,860){\makebox(0,0)[r]{ 9000}}
\put(1419.0,860.0){\rule[-0.200pt]{4.818pt}{0.400pt}}
\put(160.0,82.0){\rule[-0.200pt]{0.400pt}{4.818pt}}
\put(160,41){\makebox(0,0){ 0}}
\put(160.0,840.0){\rule[-0.200pt]{0.400pt}{4.818pt}}
\put(289.0,82.0){\rule[-0.200pt]{0.400pt}{4.818pt}}
\put(289,41){\makebox(0,0){ 10}}
\put(289.0,840.0){\rule[-0.200pt]{0.400pt}{4.818pt}}
\put(418.0,82.0){\rule[-0.200pt]{0.400pt}{4.818pt}}
\put(418,41){\makebox(0,0){ 20}}
\put(418.0,840.0){\rule[-0.200pt]{0.400pt}{4.818pt}}
\put(548.0,82.0){\rule[-0.200pt]{0.400pt}{4.818pt}}
\put(548,41){\makebox(0,0){ 30}}
\put(548.0,840.0){\rule[-0.200pt]{0.400pt}{4.818pt}}
\put(677.0,82.0){\rule[-0.200pt]{0.400pt}{4.818pt}}
\put(677,41){\makebox(0,0){ 40}}
\put(677.0,840.0){\rule[-0.200pt]{0.400pt}{4.818pt}}
\put(806.0,82.0){\rule[-0.200pt]{0.400pt}{4.818pt}}
\put(806,41){\makebox(0,0){ 50}}
\put(806.0,840.0){\rule[-0.200pt]{0.400pt}{4.818pt}}
\put(935.0,82.0){\rule[-0.200pt]{0.400pt}{4.818pt}}
\put(935,41){\makebox(0,0){ 60}}
\put(935.0,840.0){\rule[-0.200pt]{0.400pt}{4.818pt}}
\put(1064.0,82.0){\rule[-0.200pt]{0.400pt}{4.818pt}}
\put(1064,41){\makebox(0,0){ 70}}
\put(1064.0,840.0){\rule[-0.200pt]{0.400pt}{4.818pt}}
\put(1194.0,82.0){\rule[-0.200pt]{0.400pt}{4.818pt}}
\put(1194,41){\makebox(0,0){ 80}}
\put(1194.0,840.0){\rule[-0.200pt]{0.400pt}{4.818pt}}
\put(1323.0,82.0){\rule[-0.200pt]{0.400pt}{4.818pt}}
\put(1323,41){\makebox(0,0){ 90}}
\put(1323.0,840.0){\rule[-0.200pt]{0.400pt}{4.818pt}}
%\put(160.0,82.0){\rule[-0.200pt]{308.111pt}{0.400pt}}
\put(160.0,82.0){\rule[-0.200pt]{217.43pt}{0.400pt}}
%\put(1439.0,82.0){\rule[-0.200pt]{0.400pt}{187.420pt}}
\put(1439.0,82.0){\rule[-0.200pt]{0.400pt}{132.26pt}}
%\put(160.0,860.0){\rule[-0.200pt]{308.111pt}{0.400pt}}
\put(160.0,860.0){\rule[-0.200pt]{217.43pt}{0.400pt}}
%\put(160.0,860.0){\rule[-0.200pt]{215.0pt}{0.400pt}}
\put(160.0,82.0){\rule[-0.200pt]{0.400pt}{132.26pt}}
\put(160,83){\usebox{\plotpoint}}
\multiput(160.00,83.60)(1.797,0.468){5}{\rule{1.400pt}{0.113pt}}
\multiput(160.00,82.17)(10.094,4.000){2}{\rule{0.700pt}{0.400pt}}
\multiput(173.00,87.59)(1.378,0.477){7}{\rule{1.140pt}{0.115pt}}
\multiput(173.00,86.17)(10.634,5.000){2}{\rule{0.570pt}{0.400pt}}
\put(186,92.17){\rule{2.700pt}{0.400pt}}
\multiput(186.00,91.17)(7.396,2.000){2}{\rule{1.350pt}{0.400pt}}
\multiput(199.00,92.94)(1.797,-0.468){5}{\rule{1.400pt}{0.113pt}}
\multiput(199.00,93.17)(10.094,-4.000){2}{\rule{0.700pt}{0.400pt}}
\multiput(212.00,88.95)(2.695,-0.447){3}{\rule{1.833pt}{0.108pt}}
\multiput(212.00,89.17)(9.195,-3.000){2}{\rule{0.917pt}{0.400pt}}
\multiput(225.00,85.95)(2.695,-0.447){3}{\rule{1.833pt}{0.108pt}}
\multiput(225.00,86.17)(9.195,-3.000){2}{\rule{0.917pt}{0.400pt}}
\put(238,82.67){\rule{2.891pt}{0.400pt}}
\multiput(238.00,83.17)(6.000,-1.000){2}{\rule{1.445pt}{0.400pt}}
\put(289,82.67){\rule{3.132pt}{0.400pt}}
\multiput(289.00,82.17)(6.500,1.000){2}{\rule{1.566pt}{0.400pt}}
\put(250.0,83.0){\rule[-0.200pt]{9.395pt}{0.400pt}}
\put(315,84.17){\rule{2.700pt}{0.400pt}}
\multiput(315.00,83.17)(7.396,2.000){2}{\rule{1.350pt}{0.400pt}}
\put(328,86.17){\rule{2.700pt}{0.400pt}}
\multiput(328.00,85.17)(7.396,2.000){2}{\rule{1.350pt}{0.400pt}}
\multiput(341.00,88.61)(2.695,0.447){3}{\rule{1.833pt}{0.108pt}}
\multiput(341.00,87.17)(9.195,3.000){2}{\rule{0.917pt}{0.400pt}}
\multiput(354.00,91.60)(1.797,0.468){5}{\rule{1.400pt}{0.113pt}}
\multiput(354.00,90.17)(10.094,4.000){2}{\rule{0.700pt}{0.400pt}}
\put(367,93.17){\rule{2.700pt}{0.400pt}}
\multiput(367.00,94.17)(7.396,-2.000){2}{\rule{1.350pt}{0.400pt}}
\put(380,93.17){\rule{2.700pt}{0.400pt}}
\multiput(380.00,92.17)(7.396,2.000){2}{\rule{1.350pt}{0.400pt}}
\multiput(393.00,95.61)(2.472,0.447){3}{\rule{1.700pt}{0.108pt}}
\multiput(393.00,94.17)(8.472,3.000){2}{\rule{0.850pt}{0.400pt}}
\multiput(405.00,98.60)(1.797,0.468){5}{\rule{1.400pt}{0.113pt}}
\multiput(405.00,97.17)(10.094,4.000){2}{\rule{0.700pt}{0.400pt}}
\multiput(418.00,102.61)(2.695,0.447){3}{\rule{1.833pt}{0.108pt}}
\multiput(418.00,101.17)(9.195,3.000){2}{\rule{0.917pt}{0.400pt}}
\put(431,104.67){\rule{3.132pt}{0.400pt}}
\multiput(431.00,104.17)(6.500,1.000){2}{\rule{1.566pt}{0.400pt}}
\multiput(444.00,106.61)(2.695,0.447){3}{\rule{1.833pt}{0.108pt}}
\multiput(444.00,105.17)(9.195,3.000){2}{\rule{0.917pt}{0.400pt}}
\multiput(457.00,107.93)(1.123,-0.482){9}{\rule{0.967pt}{0.116pt}}
\multiput(457.00,108.17)(10.994,-6.000){2}{\rule{0.483pt}{0.400pt}}
\put(302.0,84.0){\rule[-0.200pt]{3.132pt}{0.400pt}}
\multiput(483.00,101.93)(0.728,-0.489){15}{\rule{0.678pt}{0.118pt}}
\multiput(483.00,102.17)(11.593,-9.000){2}{\rule{0.339pt}{0.400pt}}
\put(496,92.17){\rule{2.700pt}{0.400pt}}
\multiput(496.00,93.17)(7.396,-2.000){2}{\rule{1.350pt}{0.400pt}}
\multiput(509.00,90.95)(2.695,-0.447){3}{\rule{1.833pt}{0.108pt}}
\multiput(509.00,91.17)(9.195,-3.000){2}{\rule{0.917pt}{0.400pt}}
\put(522,87.17){\rule{2.700pt}{0.400pt}}
\multiput(522.00,88.17)(7.396,-2.000){2}{\rule{1.350pt}{0.400pt}}
\put(470.0,103.0){\rule[-0.200pt]{3.132pt}{0.400pt}}
\put(548,85.67){\rule{2.891pt}{0.400pt}}
\multiput(548.00,86.17)(6.000,-1.000){2}{\rule{1.445pt}{0.400pt}}
\put(535.0,87.0){\rule[-0.200pt]{3.132pt}{0.400pt}}
\put(573,86.17){\rule{2.700pt}{0.400pt}}
\multiput(573.00,85.17)(7.396,2.000){2}{\rule{1.350pt}{0.400pt}}
\put(560.0,86.0){\rule[-0.200pt]{3.132pt}{0.400pt}}
\put(612,87.67){\rule{3.132pt}{0.400pt}}
\multiput(612.00,87.17)(6.500,1.000){2}{\rule{1.566pt}{0.400pt}}
\multiput(625.00,89.61)(2.695,0.447){3}{\rule{1.833pt}{0.108pt}}
\multiput(625.00,88.17)(9.195,3.000){2}{\rule{0.917pt}{0.400pt}}
\put(638,91.67){\rule{3.132pt}{0.400pt}}
\multiput(638.00,91.17)(6.500,1.000){2}{\rule{1.566pt}{0.400pt}}
\put(651,93.17){\rule{2.700pt}{0.400pt}}
\multiput(651.00,92.17)(7.396,2.000){2}{\rule{1.350pt}{0.400pt}}
\multiput(664.00,95.61)(2.695,0.447){3}{\rule{1.833pt}{0.108pt}}
\multiput(664.00,94.17)(9.195,3.000){2}{\rule{0.917pt}{0.400pt}}
\put(586.0,88.0){\rule[-0.200pt]{6.263pt}{0.400pt}}
\put(690,98.17){\rule{2.700pt}{0.400pt}}
\multiput(690.00,97.17)(7.396,2.000){2}{\rule{1.350pt}{0.400pt}}
\put(677.0,98.0){\rule[-0.200pt]{3.132pt}{0.400pt}}
\multiput(716.00,100.61)(2.472,0.447){3}{\rule{1.700pt}{0.108pt}}
\multiput(716.00,99.17)(8.472,3.000){2}{\rule{0.850pt}{0.400pt}}
\put(728,102.67){\rule{3.132pt}{0.400pt}}
\multiput(728.00,102.17)(6.500,1.000){2}{\rule{1.566pt}{0.400pt}}
\multiput(741.00,104.59)(1.378,0.477){7}{\rule{1.140pt}{0.115pt}}
\multiput(741.00,103.17)(10.634,5.000){2}{\rule{0.570pt}{0.400pt}}
\multiput(754.00,109.60)(1.797,0.468){5}{\rule{1.400pt}{0.113pt}}
\multiput(754.00,108.17)(10.094,4.000){2}{\rule{0.700pt}{0.400pt}}
\put(767,112.67){\rule{3.132pt}{0.400pt}}
\multiput(767.00,112.17)(6.500,1.000){2}{\rule{1.566pt}{0.400pt}}
\multiput(780.00,114.61)(2.695,0.447){3}{\rule{1.833pt}{0.108pt}}
\multiput(780.00,113.17)(9.195,3.000){2}{\rule{0.917pt}{0.400pt}}
\multiput(793.58,117.00)(0.493,0.695){23}{\rule{0.119pt}{0.654pt}}
\multiput(792.17,117.00)(13.000,16.643){2}{\rule{0.400pt}{0.327pt}}
\multiput(806.00,133.93)(1.378,-0.477){7}{\rule{1.140pt}{0.115pt}}
\multiput(806.00,134.17)(10.634,-5.000){2}{\rule{0.570pt}{0.400pt}}
\put(819,128.67){\rule{3.132pt}{0.400pt}}
\multiput(819.00,129.17)(6.500,-1.000){2}{\rule{1.566pt}{0.400pt}}
\multiput(832.00,129.59)(0.824,0.488){13}{\rule{0.750pt}{0.117pt}}
\multiput(832.00,128.17)(11.443,8.000){2}{\rule{0.375pt}{0.400pt}}
\multiput(845.00,135.93)(0.728,-0.489){15}{\rule{0.678pt}{0.118pt}}
\multiput(845.00,136.17)(11.593,-9.000){2}{\rule{0.339pt}{0.400pt}}
\put(858,128.17){\rule{2.700pt}{0.400pt}}
\multiput(858.00,127.17)(7.396,2.000){2}{\rule{1.350pt}{0.400pt}}
\put(871,128.17){\rule{2.500pt}{0.400pt}}
\multiput(871.00,129.17)(6.811,-2.000){2}{\rule{1.250pt}{0.400pt}}
\put(883,128.17){\rule{2.700pt}{0.400pt}}
\multiput(883.00,127.17)(7.396,2.000){2}{\rule{1.350pt}{0.400pt}}
\multiput(896.00,130.58)(0.652,0.491){17}{\rule{0.620pt}{0.118pt}}
\multiput(896.00,129.17)(11.713,10.000){2}{\rule{0.310pt}{0.400pt}}
\multiput(909.58,137.29)(0.493,-0.695){23}{\rule{0.119pt}{0.654pt}}
\multiput(908.17,138.64)(13.000,-16.643){2}{\rule{0.400pt}{0.327pt}}
\put(922,120.67){\rule{3.132pt}{0.400pt}}
\multiput(922.00,121.17)(6.500,-1.000){2}{\rule{1.566pt}{0.400pt}}
\put(935,120.67){\rule{3.132pt}{0.400pt}}
\multiput(935.00,120.17)(6.500,1.000){2}{\rule{1.566pt}{0.400pt}}
\multiput(948.00,120.93)(1.378,-0.477){7}{\rule{1.140pt}{0.115pt}}
\multiput(948.00,121.17)(10.634,-5.000){2}{\rule{0.570pt}{0.400pt}}
\put(961,115.67){\rule{3.132pt}{0.400pt}}
\multiput(961.00,116.17)(6.500,-1.000){2}{\rule{1.566pt}{0.400pt}}
\multiput(974.00,114.94)(1.797,-0.468){5}{\rule{1.400pt}{0.113pt}}
\multiput(974.00,115.17)(10.094,-4.000){2}{\rule{0.700pt}{0.400pt}}
\multiput(987.00,110.95)(2.695,-0.447){3}{\rule{1.833pt}{0.108pt}}
\multiput(987.00,111.17)(9.195,-3.000){2}{\rule{0.917pt}{0.400pt}}
\put(1000,109.17){\rule{2.700pt}{0.400pt}}
\multiput(1000.00,108.17)(7.396,2.000){2}{\rule{1.350pt}{0.400pt}}
\multiput(1013.00,109.93)(0.728,-0.489){15}{\rule{0.678pt}{0.118pt}}
\multiput(1013.00,110.17)(11.593,-9.000){2}{\rule{0.339pt}{0.400pt}}
\put(703.0,100.0){\rule[-0.200pt]{3.132pt}{0.400pt}}
\multiput(1051.00,100.95)(2.695,-0.447){3}{\rule{1.833pt}{0.108pt}}
\multiput(1051.00,101.17)(9.195,-3.000){2}{\rule{0.917pt}{0.400pt}}
\put(1064,97.17){\rule{2.700pt}{0.400pt}}
\multiput(1064.00,98.17)(7.396,-2.000){2}{\rule{1.350pt}{0.400pt}}
\multiput(1077.00,95.95)(2.695,-0.447){3}{\rule{1.833pt}{0.108pt}}
\multiput(1077.00,96.17)(9.195,-3.000){2}{\rule{0.917pt}{0.400pt}}
\put(1090,93.67){\rule{3.132pt}{0.400pt}}
\multiput(1090.00,93.17)(6.500,1.000){2}{\rule{1.566pt}{0.400pt}}
\put(1103,93.17){\rule{2.700pt}{0.400pt}}
\multiput(1103.00,94.17)(7.396,-2.000){2}{\rule{1.350pt}{0.400pt}}
\multiput(1116.00,93.61)(2.695,0.447){3}{\rule{1.833pt}{0.108pt}}
\multiput(1116.00,92.17)(9.195,3.000){2}{\rule{0.917pt}{0.400pt}}
\put(1129,96.17){\rule{2.700pt}{0.400pt}}
\multiput(1129.00,95.17)(7.396,2.000){2}{\rule{1.350pt}{0.400pt}}
\put(1142,96.67){\rule{3.132pt}{0.400pt}}
\multiput(1142.00,97.17)(6.500,-1.000){2}{\rule{1.566pt}{0.400pt}}
\put(1026.0,102.0){\rule[-0.200pt]{6.022pt}{0.400pt}}
\multiput(1168.00,97.61)(2.695,0.447){3}{\rule{1.833pt}{0.108pt}}
\multiput(1168.00,96.17)(9.195,3.000){2}{\rule{0.917pt}{0.400pt}}
\put(1181,98.67){\rule{3.132pt}{0.400pt}}
\multiput(1181.00,99.17)(6.500,-1.000){2}{\rule{1.566pt}{0.400pt}}
\multiput(1194.00,99.60)(1.651,0.468){5}{\rule{1.300pt}{0.113pt}}
\multiput(1194.00,98.17)(9.302,4.000){2}{\rule{0.650pt}{0.400pt}}
\put(1206,101.17){\rule{2.700pt}{0.400pt}}
\multiput(1206.00,102.17)(7.396,-2.000){2}{\rule{1.350pt}{0.400pt}}
\multiput(1219.00,101.58)(0.652,0.491){17}{\rule{0.620pt}{0.118pt}}
\multiput(1219.00,100.17)(11.713,10.000){2}{\rule{0.310pt}{0.400pt}}
\put(1232,109.17){\rule{2.700pt}{0.400pt}}
\multiput(1232.00,110.17)(7.396,-2.000){2}{\rule{1.350pt}{0.400pt}}
\multiput(1245.00,109.58)(0.539,0.492){21}{\rule{0.533pt}{0.119pt}}
\multiput(1245.00,108.17)(11.893,12.000){2}{\rule{0.267pt}{0.400pt}}
\multiput(1258.00,121.59)(0.728,0.489){15}{\rule{0.678pt}{0.118pt}}
\multiput(1258.00,120.17)(11.593,9.000){2}{\rule{0.339pt}{0.400pt}}
\multiput(1271.58,130.00)(0.493,0.576){23}{\rule{0.119pt}{0.562pt}}
\multiput(1270.17,130.00)(13.000,13.834){2}{\rule{0.400pt}{0.281pt}}
\multiput(1284.58,145.00)(0.493,0.655){23}{\rule{0.119pt}{0.623pt}}
\multiput(1283.17,145.00)(13.000,15.707){2}{\rule{0.400pt}{0.312pt}}
\multiput(1297.00,162.58)(0.652,0.491){17}{\rule{0.620pt}{0.118pt}}
\multiput(1297.00,161.17)(11.713,10.000){2}{\rule{0.310pt}{0.400pt}}
\multiput(1310.00,172.58)(0.590,0.492){19}{\rule{0.573pt}{0.118pt}}
\multiput(1310.00,171.17)(11.811,11.000){2}{\rule{0.286pt}{0.400pt}}
\multiput(1323.58,183.00)(0.493,1.012){23}{\rule{0.119pt}{0.900pt}}
\multiput(1322.17,183.00)(13.000,24.132){2}{\rule{0.400pt}{0.450pt}}
\multiput(1336.58,206.67)(0.493,-0.576){23}{\rule{0.119pt}{0.562pt}}
\multiput(1335.17,207.83)(13.000,-13.834){2}{\rule{0.400pt}{0.281pt}}
\multiput(1349.00,194.61)(2.472,0.447){3}{\rule{1.700pt}{0.108pt}}
\multiput(1349.00,193.17)(8.472,3.000){2}{\rule{0.850pt}{0.400pt}}
\multiput(1361.58,197.00)(0.493,0.695){23}{\rule{0.119pt}{0.654pt}}
\multiput(1360.17,197.00)(13.000,16.643){2}{\rule{0.400pt}{0.327pt}}
\multiput(1374.58,215.00)(0.493,1.805){23}{\rule{0.119pt}{1.515pt}}
\multiput(1373.17,215.00)(13.000,42.855){2}{\rule{0.400pt}{0.758pt}}
\multiput(1387.58,261.00)(0.493,1.924){23}{\rule{0.119pt}{1.608pt}}
\multiput(1386.17,261.00)(13.000,45.663){2}{\rule{0.400pt}{0.804pt}}
\multiput(1400.58,310.00)(0.493,2.083){23}{\rule{0.119pt}{1.731pt}}
\multiput(1399.17,310.00)(13.000,49.408){2}{\rule{0.400pt}{0.865pt}}
\multiput(1413.58,363.00)(0.493,3.391){23}{\rule{0.119pt}{2.746pt}}
\multiput(1412.17,363.00)(13.000,80.300){2}{\rule{0.400pt}{1.373pt}}
\multiput(1426.58,449.00)(0.493,13.977){23}{\rule{0.119pt}{10.962pt}}
\multiput(1425.17,449.00)(13.000,330.249){2}{\rule{0.400pt}{5.481pt}}
\put(1155.0,97.0){\rule[-0.200pt]{3.132pt}{0.400pt}}
\put(1439,802){\usebox{\plotpoint}}
%\put(160.0,82.0){\rule[-0.200pt]{308.111pt}{0.400pt}}
%\put(1439.0,82.0){\rule[-0.200pt]{0.400pt}{187.420pt}}
%\put(160.0,860.0){\rule[-0.200pt]{308.111pt}{0.400pt}}
%\put(160.0,82.0){\rule[-0.200pt]{0.400pt}{187.420pt}}
\end{picture}

%% file: stat9blog-50.tex
% GNUPLOT: LaTeX picture
%\setlength{\unitlength}{0.240900pt}
\setlength{\unitlength}{0.168pt}
\ifx\plotpoint\undefined\newsavebox{\plotpoint}\fi
\sbox{\plotpoint}{\rule[-0.200pt]{0.400pt}{0.400pt}}%
\begin{picture}(1500,900)(0,0)
\sbox{\plotpoint}{\rule[-0.200pt]{0.400pt}{0.400pt}}%
\put(180.0,82.0){\rule[-0.200pt]{4.818pt}{0.400pt}}
\put(160,82){\makebox(0,0)[r]{ 10}}
\put(1419.0,82.0){\rule[-0.200pt]{4.818pt}{0.400pt}}
\put(180.0,160.0){\rule[-0.200pt]{2.409pt}{0.400pt}}
\put(1429.0,160.0){\rule[-0.200pt]{2.409pt}{0.400pt}}
\put(180.0,206.0){\rule[-0.200pt]{2.409pt}{0.400pt}}
\put(1429.0,206.0){\rule[-0.200pt]{2.409pt}{0.400pt}}
\put(180.0,238.0){\rule[-0.200pt]{2.409pt}{0.400pt}}
\put(1429.0,238.0){\rule[-0.200pt]{2.409pt}{0.400pt}}
\put(180.0,263.0){\rule[-0.200pt]{2.409pt}{0.400pt}}
\put(1429.0,263.0){\rule[-0.200pt]{2.409pt}{0.400pt}}
\put(180.0,284.0){\rule[-0.200pt]{2.409pt}{0.400pt}}
\put(1429.0,284.0){\rule[-0.200pt]{2.409pt}{0.400pt}}
\put(180.0,301.0){\rule[-0.200pt]{2.409pt}{0.400pt}}
\put(1429.0,301.0){\rule[-0.200pt]{2.409pt}{0.400pt}}
\put(180.0,316.0){\rule[-0.200pt]{2.409pt}{0.400pt}}
\put(1429.0,316.0){\rule[-0.200pt]{2.409pt}{0.400pt}}
\put(180.0,329.0){\rule[-0.200pt]{2.409pt}{0.400pt}}
\put(1429.0,329.0){\rule[-0.200pt]{2.409pt}{0.400pt}}
\put(180.0,341.0){\rule[-0.200pt]{4.818pt}{0.400pt}}
\put(160,341){\makebox(0,0)[r]{ 100}}
\put(1419.0,341.0){\rule[-0.200pt]{4.818pt}{0.400pt}}
\put(180.0,419.0){\rule[-0.200pt]{2.409pt}{0.400pt}}
\put(1429.0,419.0){\rule[-0.200pt]{2.409pt}{0.400pt}}
\put(180.0,465.0){\rule[-0.200pt]{2.409pt}{0.400pt}}
\put(1429.0,465.0){\rule[-0.200pt]{2.409pt}{0.400pt}}
\put(180.0,497.0){\rule[-0.200pt]{2.409pt}{0.400pt}}
\put(1429.0,497.0){\rule[-0.200pt]{2.409pt}{0.400pt}}
\put(180.0,523.0){\rule[-0.200pt]{2.409pt}{0.400pt}}
\put(1429.0,523.0){\rule[-0.200pt]{2.409pt}{0.400pt}}
\put(180.0,543.0){\rule[-0.200pt]{2.409pt}{0.400pt}}
\put(1429.0,543.0){\rule[-0.200pt]{2.409pt}{0.400pt}}
\put(180.0,560.0){\rule[-0.200pt]{2.409pt}{0.400pt}}
\put(1429.0,560.0){\rule[-0.200pt]{2.409pt}{0.400pt}}
\put(180.0,576.0){\rule[-0.200pt]{2.409pt}{0.400pt}}
\put(1429.0,576.0){\rule[-0.200pt]{2.409pt}{0.400pt}}
\put(180.0,589.0){\rule[-0.200pt]{2.409pt}{0.400pt}}
\put(1429.0,589.0){\rule[-0.200pt]{2.409pt}{0.400pt}}
\put(180.0,601.0){\rule[-0.200pt]{4.818pt}{0.400pt}}
\put(160,601){\makebox(0,0)[r]{ 1000}}
\put(1419.0,601.0){\rule[-0.200pt]{4.818pt}{0.400pt}}
\put(180.0,679.0){\rule[-0.200pt]{2.409pt}{0.400pt}}
\put(1429.0,679.0){\rule[-0.200pt]{2.409pt}{0.400pt}}
\put(180.0,724.0){\rule[-0.200pt]{2.409pt}{0.400pt}}
\put(1429.0,724.0){\rule[-0.200pt]{2.409pt}{0.400pt}}
\put(180.0,757.0){\rule[-0.200pt]{2.409pt}{0.400pt}}
\put(1429.0,757.0){\rule[-0.200pt]{2.409pt}{0.400pt}}
\put(180.0,782.0){\rule[-0.200pt]{2.409pt}{0.400pt}}
\put(1429.0,782.0){\rule[-0.200pt]{2.409pt}{0.400pt}}
\put(180.0,802.0){\rule[-0.200pt]{2.409pt}{0.400pt}}
\put(1429.0,802.0){\rule[-0.200pt]{2.409pt}{0.400pt}}
\put(180.0,820.0){\rule[-0.200pt]{2.409pt}{0.400pt}}
\put(1429.0,820.0){\rule[-0.200pt]{2.409pt}{0.400pt}}
\put(180.0,835.0){\rule[-0.200pt]{2.409pt}{0.400pt}}
\put(1429.0,835.0){\rule[-0.200pt]{2.409pt}{0.400pt}}
\put(180.0,848.0){\rule[-0.200pt]{2.409pt}{0.400pt}}
\put(1429.0,848.0){\rule[-0.200pt]{2.409pt}{0.400pt}}
\put(180.0,860.0){\rule[-0.200pt]{4.818pt}{0.400pt}}
\put(160,860){\makebox(0,0)[r]{ 10000}}
\put(1419.0,860.0){\rule[-0.200pt]{4.818pt}{0.400pt}}
\put(180.0,82.0){\rule[-0.200pt]{0.400pt}{4.818pt}}
\put(180,41){\makebox(0,0){ 0}}
\put(180.0,840.0){\rule[-0.200pt]{0.400pt}{4.818pt}}
\put(307.0,82.0){\rule[-0.200pt]{0.400pt}{4.818pt}}
\put(307,41){\makebox(0,0){ 10}}
\put(307.0,840.0){\rule[-0.200pt]{0.400pt}{4.818pt}}
\put(434.0,82.0){\rule[-0.200pt]{0.400pt}{4.818pt}}
\put(434,41){\makebox(0,0){ 20}}
\put(434.0,840.0){\rule[-0.200pt]{0.400pt}{4.818pt}}
\put(562.0,82.0){\rule[-0.200pt]{0.400pt}{4.818pt}}
\put(562,41){\makebox(0,0){ 30}}
\put(562.0,840.0){\rule[-0.200pt]{0.400pt}{4.818pt}}
\put(689.0,82.0){\rule[-0.200pt]{0.400pt}{4.818pt}}
\put(689,41){\makebox(0,0){ 40}}
\put(689.0,840.0){\rule[-0.200pt]{0.400pt}{4.818pt}}
\put(816.0,82.0){\rule[-0.200pt]{0.400pt}{4.818pt}}
\put(816,41){\makebox(0,0){ 50}}
\put(816.0,840.0){\rule[-0.200pt]{0.400pt}{4.818pt}}
\put(943.0,82.0){\rule[-0.200pt]{0.400pt}{4.818pt}}
\put(943,41){\makebox(0,0){ 60}}
\put(943.0,840.0){\rule[-0.200pt]{0.400pt}{4.818pt}}
\put(1070.0,82.0){\rule[-0.200pt]{0.400pt}{4.818pt}}
\put(1070,41){\makebox(0,0){ 70}}
\put(1070.0,840.0){\rule[-0.200pt]{0.400pt}{4.818pt}}
\put(1197.0,82.0){\rule[-0.200pt]{0.400pt}{4.818pt}}
\put(1197,41){\makebox(0,0){ 80}}
\put(1197.0,840.0){\rule[-0.200pt]{0.400pt}{4.818pt}}
\put(1325.0,82.0){\rule[-0.200pt]{0.400pt}{4.818pt}}
\put(1325,41){\makebox(0,0){ 90}}
\put(1325.0,840.0){\rule[-0.200pt]{0.400pt}{4.818pt}}
%\put(180.0,82.0){\rule[-0.200pt]{303.293pt}{0.400pt}}
\put(180.0,82.0){\rule[-0.200pt]{214.03pt}{0.400pt}}
%\put(1439.0,82.0){\rule[-0.200pt]{0.400pt}{187.420pt}}
\put(1439.0,82.0){\rule[-0.200pt]{0.400pt}{132.26pt}}
%\put(180.0,860.0){\rule[-0.200pt]{303.293pt}{0.400pt}}
\put(180.0,860.0){\rule[-0.200pt]{214.03pt}{0.400pt}}
%\put(180.0,82.0){\rule[-0.200pt]{0.400pt}{187.420pt}}
\put(180.0,82.0){\rule[-0.200pt]{0.400pt}{132.26pt}}
\put(180,93){\usebox{\plotpoint}}
\multiput(180.58,93.00)(0.493,6.999){23}{\rule{0.119pt}{5.546pt}}
\multiput(179.17,93.00)(13.000,165.489){2}{\rule{0.400pt}{2.773pt}}
\multiput(193.58,270.00)(0.492,3.857){21}{\rule{0.119pt}{3.100pt}}
\multiput(192.17,270.00)(12.000,83.566){2}{\rule{0.400pt}{1.550pt}}
\multiput(205.58,360.00)(0.493,0.616){23}{\rule{0.119pt}{0.592pt}}
\multiput(204.17,360.00)(13.000,14.771){2}{\rule{0.400pt}{0.296pt}}
\multiput(218.58,369.58)(0.493,-1.845){23}{\rule{0.119pt}{1.546pt}}
\multiput(217.17,372.79)(13.000,-43.791){2}{\rule{0.400pt}{0.773pt}}
\multiput(231.58,323.09)(0.493,-1.686){23}{\rule{0.119pt}{1.423pt}}
\multiput(230.17,326.05)(13.000,-40.046){2}{\rule{0.400pt}{0.712pt}}
\multiput(244.58,269.67)(0.492,-4.934){21}{\rule{0.119pt}{3.933pt}}
\multiput(243.17,277.84)(12.000,-106.836){2}{\rule{0.400pt}{1.967pt}}
\multiput(256.58,164.07)(0.493,-2.003){23}{\rule{0.119pt}{1.669pt}}
\multiput(255.17,167.54)(13.000,-47.535){2}{\rule{0.400pt}{0.835pt}}
\multiput(269.58,117.41)(0.493,-0.655){23}{\rule{0.119pt}{0.623pt}}
\multiput(268.17,118.71)(13.000,-15.707){2}{\rule{0.400pt}{0.312pt}}
\multiput(282.58,99.68)(0.492,-0.884){21}{\rule{0.119pt}{0.800pt}}
\multiput(281.17,101.34)(12.000,-19.340){2}{\rule{0.400pt}{0.400pt}}
\multiput(294.58,82.00)(0.493,1.171){23}{\rule{0.119pt}{1.023pt}}
\multiput(293.17,82.00)(13.000,27.877){2}{\rule{0.400pt}{0.512pt}}
\multiput(307.58,112.00)(0.493,1.646){23}{\rule{0.119pt}{1.392pt}}
\multiput(306.17,112.00)(13.000,39.110){2}{\rule{0.400pt}{0.696pt}}
\multiput(320.58,154.00)(0.493,1.726){23}{\rule{0.119pt}{1.454pt}}
\multiput(319.17,154.00)(13.000,40.982){2}{\rule{0.400pt}{0.727pt}}
\multiput(333.58,198.00)(0.492,2.047){21}{\rule{0.119pt}{1.700pt}}
\multiput(332.17,198.00)(12.000,44.472){2}{\rule{0.400pt}{0.850pt}}
\multiput(345.58,246.00)(0.493,2.122){23}{\rule{0.119pt}{1.762pt}}
\multiput(344.17,246.00)(13.000,50.344){2}{\rule{0.400pt}{0.881pt}}
\multiput(358.58,300.00)(0.493,1.646){23}{\rule{0.119pt}{1.392pt}}
\multiput(357.17,300.00)(13.000,39.110){2}{\rule{0.400pt}{0.696pt}}
\multiput(371.58,342.00)(0.492,1.789){21}{\rule{0.119pt}{1.500pt}}
\multiput(370.17,342.00)(12.000,38.887){2}{\rule{0.400pt}{0.750pt}}
\multiput(383.58,381.41)(0.493,-0.655){23}{\rule{0.119pt}{0.623pt}}
\multiput(382.17,382.71)(13.000,-15.707){2}{\rule{0.400pt}{0.312pt}}
\multiput(396.58,367.00)(0.493,0.814){23}{\rule{0.119pt}{0.746pt}}
\multiput(395.17,367.00)(13.000,19.451){2}{\rule{0.400pt}{0.373pt}}
\multiput(409.58,388.00)(0.493,0.853){23}{\rule{0.119pt}{0.777pt}}
\multiput(408.17,388.00)(13.000,20.387){2}{\rule{0.400pt}{0.388pt}}
\multiput(422.58,410.00)(0.492,1.056){21}{\rule{0.119pt}{0.933pt}}
\multiput(421.17,410.00)(12.000,23.063){2}{\rule{0.400pt}{0.467pt}}
\multiput(434.58,435.00)(0.493,0.536){23}{\rule{0.119pt}{0.531pt}}
\multiput(433.17,435.00)(13.000,12.898){2}{\rule{0.400pt}{0.265pt}}
\multiput(447.00,449.59)(0.950,0.485){11}{\rule{0.843pt}{0.117pt}}
\multiput(447.00,448.17)(11.251,7.000){2}{\rule{0.421pt}{0.400pt}}
\multiput(460.58,456.00)(0.492,0.539){21}{\rule{0.119pt}{0.533pt}}
\multiput(459.17,456.00)(12.000,11.893){2}{\rule{0.400pt}{0.267pt}}
\multiput(472.58,464.88)(0.493,-1.131){23}{\rule{0.119pt}{0.992pt}}
\multiput(471.17,466.94)(13.000,-26.940){2}{\rule{0.400pt}{0.496pt}}
\multiput(498.58,431.67)(0.493,-2.439){23}{\rule{0.119pt}{2.008pt}}
\multiput(497.17,435.83)(13.000,-57.833){2}{\rule{0.400pt}{1.004pt}}
\multiput(511.58,374.96)(0.492,-0.798){21}{\rule{0.119pt}{0.733pt}}
\multiput(510.17,376.48)(12.000,-17.478){2}{\rule{0.400pt}{0.367pt}}
\multiput(523.58,352.33)(0.493,-1.924){23}{\rule{0.119pt}{1.608pt}}
\multiput(522.17,355.66)(13.000,-45.663){2}{\rule{0.400pt}{0.804pt}}
\multiput(536.58,306.52)(0.493,-0.933){23}{\rule{0.119pt}{0.838pt}}
\multiput(535.17,308.26)(13.000,-22.260){2}{\rule{0.400pt}{0.419pt}}
\multiput(549.00,284.94)(1.797,-0.468){5}{\rule{1.400pt}{0.113pt}}
\multiput(549.00,285.17)(10.094,-4.000){2}{\rule{0.700pt}{0.400pt}}
\multiput(562.58,277.71)(0.492,-1.186){21}{\rule{0.119pt}{1.033pt}}
\multiput(561.17,279.86)(12.000,-25.855){2}{\rule{0.400pt}{0.517pt}}
\put(574,254.17){\rule{2.700pt}{0.400pt}}
\multiput(574.00,253.17)(7.396,2.000){2}{\rule{1.350pt}{0.400pt}}
\multiput(587.58,256.00)(0.493,1.567){23}{\rule{0.119pt}{1.331pt}}
\multiput(586.17,256.00)(13.000,37.238){2}{\rule{0.400pt}{0.665pt}}
\multiput(600.00,296.60)(1.651,0.468){5}{\rule{1.300pt}{0.113pt}}
\multiput(600.00,295.17)(9.302,4.000){2}{\rule{0.650pt}{0.400pt}}
\multiput(612.00,300.59)(1.123,0.482){9}{\rule{0.967pt}{0.116pt}}
\multiput(612.00,299.17)(10.994,6.000){2}{\rule{0.483pt}{0.400pt}}
\multiput(625.00,306.58)(0.652,0.491){17}{\rule{0.620pt}{0.118pt}}
\multiput(625.00,305.17)(11.713,10.000){2}{\rule{0.310pt}{0.400pt}}
\multiput(638.58,316.00)(0.493,1.845){23}{\rule{0.119pt}{1.546pt}}
\multiput(637.17,316.00)(13.000,43.791){2}{\rule{0.400pt}{0.773pt}}
\multiput(651.00,363.59)(1.033,0.482){9}{\rule{0.900pt}{0.116pt}}
\multiput(651.00,362.17)(10.132,6.000){2}{\rule{0.450pt}{0.400pt}}
\multiput(663.58,369.00)(0.493,0.695){23}{\rule{0.119pt}{0.654pt}}
\multiput(662.17,369.00)(13.000,16.643){2}{\rule{0.400pt}{0.327pt}}
\multiput(676.58,387.00)(0.493,0.814){23}{\rule{0.119pt}{0.746pt}}
\multiput(675.17,387.00)(13.000,19.451){2}{\rule{0.400pt}{0.373pt}}
\multiput(689.00,408.61)(2.472,0.447){3}{\rule{1.700pt}{0.108pt}}
\multiput(689.00,407.17)(8.472,3.000){2}{\rule{0.850pt}{0.400pt}}
\multiput(701.00,411.58)(0.652,0.491){17}{\rule{0.620pt}{0.118pt}}
\multiput(701.00,410.17)(11.713,10.000){2}{\rule{0.310pt}{0.400pt}}
\multiput(714.00,421.61)(2.695,0.447){3}{\rule{1.833pt}{0.108pt}}
\multiput(714.00,420.17)(9.195,3.000){2}{\rule{0.917pt}{0.400pt}}
\multiput(727.58,424.00)(0.493,0.616){23}{\rule{0.119pt}{0.592pt}}
\multiput(726.17,424.00)(13.000,14.771){2}{\rule{0.400pt}{0.296pt}}
\multiput(740.00,440.59)(1.267,0.477){7}{\rule{1.060pt}{0.115pt}}
\multiput(740.00,439.17)(9.800,5.000){2}{\rule{0.530pt}{0.400pt}}
\multiput(752.58,445.00)(0.493,0.893){23}{\rule{0.119pt}{0.808pt}}
\multiput(751.17,445.00)(13.000,21.324){2}{\rule{0.400pt}{0.404pt}}
\multiput(765.58,468.00)(0.493,0.695){23}{\rule{0.119pt}{0.654pt}}
\multiput(764.17,468.00)(13.000,16.643){2}{\rule{0.400pt}{0.327pt}}
\multiput(778.00,486.60)(1.651,0.468){5}{\rule{1.300pt}{0.113pt}}
\multiput(778.00,485.17)(9.302,4.000){2}{\rule{0.650pt}{0.400pt}}
\multiput(790.00,490.59)(0.728,0.489){15}{\rule{0.678pt}{0.118pt}}
\multiput(790.00,489.17)(11.593,9.000){2}{\rule{0.339pt}{0.400pt}}
\multiput(803.58,499.00)(0.493,1.805){23}{\rule{0.119pt}{1.515pt}}
\multiput(802.17,499.00)(13.000,42.855){2}{\rule{0.400pt}{0.758pt}}
\multiput(816.00,543.92)(0.652,-0.491){17}{\rule{0.620pt}{0.118pt}}
\multiput(816.00,544.17)(11.713,-10.000){2}{\rule{0.310pt}{0.400pt}}
\put(829,533.17){\rule{2.500pt}{0.400pt}}
\multiput(829.00,534.17)(6.811,-2.000){2}{\rule{1.250pt}{0.400pt}}
\multiput(841.58,533.00)(0.493,0.616){23}{\rule{0.119pt}{0.592pt}}
\multiput(840.17,533.00)(13.000,14.771){2}{\rule{0.400pt}{0.296pt}}
\multiput(854.58,546.16)(0.493,-0.734){23}{\rule{0.119pt}{0.685pt}}
\multiput(853.17,547.58)(13.000,-17.579){2}{\rule{0.400pt}{0.342pt}}
\multiput(867.00,530.60)(1.651,0.468){5}{\rule{1.300pt}{0.113pt}}
\multiput(867.00,529.17)(9.302,4.000){2}{\rule{0.650pt}{0.400pt}}
\multiput(879.00,532.93)(1.378,-0.477){7}{\rule{1.140pt}{0.115pt}}
\multiput(879.00,533.17)(10.634,-5.000){2}{\rule{0.570pt}{0.400pt}}
\multiput(892.00,529.59)(1.123,0.482){9}{\rule{0.967pt}{0.116pt}}
\multiput(892.00,528.17)(10.994,6.000){2}{\rule{0.483pt}{0.400pt}}
\multiput(905.58,535.00)(0.493,0.814){23}{\rule{0.119pt}{0.746pt}}
\multiput(904.17,535.00)(13.000,19.451){2}{\rule{0.400pt}{0.373pt}}
\multiput(918.58,549.77)(0.492,-1.789){21}{\rule{0.119pt}{1.500pt}}
\multiput(917.17,552.89)(12.000,-38.887){2}{\rule{0.400pt}{0.750pt}}
\multiput(930.00,512.94)(1.797,-0.468){5}{\rule{1.400pt}{0.113pt}}
\multiput(930.00,513.17)(10.094,-4.000){2}{\rule{0.700pt}{0.400pt}}
\multiput(943.00,510.61)(2.695,0.447){3}{\rule{1.833pt}{0.108pt}}
\multiput(943.00,509.17)(9.195,3.000){2}{\rule{0.917pt}{0.400pt}}
\multiput(956.58,510.37)(0.492,-0.669){21}{\rule{0.119pt}{0.633pt}}
\multiput(955.17,511.69)(12.000,-14.685){2}{\rule{0.400pt}{0.317pt}}
\multiput(968.00,495.95)(2.695,-0.447){3}{\rule{1.833pt}{0.108pt}}
\multiput(968.00,496.17)(9.195,-3.000){2}{\rule{0.917pt}{0.400pt}}
\multiput(981.00,492.92)(0.590,-0.492){19}{\rule{0.573pt}{0.118pt}}
\multiput(981.00,493.17)(11.811,-11.000){2}{\rule{0.286pt}{0.400pt}}
\multiput(994.58,480.67)(0.493,-0.576){23}{\rule{0.119pt}{0.562pt}}
\multiput(993.17,481.83)(13.000,-13.834){2}{\rule{0.400pt}{0.281pt}}
\multiput(1007.00,468.58)(0.543,0.492){19}{\rule{0.536pt}{0.118pt}}
\multiput(1007.00,467.17)(10.887,11.000){2}{\rule{0.268pt}{0.400pt}}
\multiput(1019.58,472.96)(0.493,-1.726){23}{\rule{0.119pt}{1.454pt}}
\multiput(1018.17,475.98)(13.000,-40.982){2}{\rule{0.400pt}{0.727pt}}
\put(485.0,440.0){\rule[-0.200pt]{3.132pt}{0.400pt}}
\multiput(1045.00,435.61)(2.472,0.447){3}{\rule{1.700pt}{0.108pt}}
\multiput(1045.00,434.17)(8.472,3.000){2}{\rule{0.850pt}{0.400pt}}
\multiput(1057.58,434.65)(0.493,-0.893){23}{\rule{0.119pt}{0.808pt}}
\multiput(1056.17,436.32)(13.000,-21.324){2}{\rule{0.400pt}{0.404pt}}
\multiput(1070.00,413.93)(0.824,-0.488){13}{\rule{0.750pt}{0.117pt}}
\multiput(1070.00,414.17)(11.443,-8.000){2}{\rule{0.375pt}{0.400pt}}
\multiput(1083.58,403.39)(0.493,-0.972){23}{\rule{0.119pt}{0.869pt}}
\multiput(1082.17,405.20)(13.000,-23.196){2}{\rule{0.400pt}{0.435pt}}
\multiput(1096.00,382.59)(1.267,0.477){7}{\rule{1.060pt}{0.115pt}}
\multiput(1096.00,381.17)(9.800,5.000){2}{\rule{0.530pt}{0.400pt}}
\multiput(1108.58,383.90)(0.493,-0.814){23}{\rule{0.119pt}{0.746pt}}
\multiput(1107.17,385.45)(13.000,-19.451){2}{\rule{0.400pt}{0.373pt}}
\multiput(1121.58,366.00)(0.493,1.250){23}{\rule{0.119pt}{1.085pt}}
\multiput(1120.17,366.00)(13.000,29.749){2}{\rule{0.400pt}{0.542pt}}
\multiput(1134.00,398.58)(0.590,0.492){19}{\rule{0.573pt}{0.118pt}}
\multiput(1134.00,397.17)(11.811,11.000){2}{\rule{0.286pt}{0.400pt}}
\multiput(1147.00,407.93)(0.874,-0.485){11}{\rule{0.786pt}{0.117pt}}
\multiput(1147.00,408.17)(10.369,-7.000){2}{\rule{0.393pt}{0.400pt}}
\multiput(1159.00,402.61)(2.695,0.447){3}{\rule{1.833pt}{0.108pt}}
\multiput(1159.00,401.17)(9.195,3.000){2}{\rule{0.917pt}{0.400pt}}
\multiput(1172.58,405.00)(0.493,0.655){23}{\rule{0.119pt}{0.623pt}}
\multiput(1171.17,405.00)(13.000,15.707){2}{\rule{0.400pt}{0.312pt}}
\multiput(1185.00,420.93)(0.874,-0.485){11}{\rule{0.786pt}{0.117pt}}
\multiput(1185.00,421.17)(10.369,-7.000){2}{\rule{0.393pt}{0.400pt}}
\multiput(1197.58,415.00)(0.493,0.972){23}{\rule{0.119pt}{0.869pt}}
\multiput(1196.17,415.00)(13.000,23.196){2}{\rule{0.400pt}{0.435pt}}
\multiput(1210.00,438.92)(0.590,-0.492){19}{\rule{0.573pt}{0.118pt}}
\multiput(1210.00,439.17)(11.811,-11.000){2}{\rule{0.286pt}{0.400pt}}
\multiput(1223.58,429.00)(0.493,1.964){23}{\rule{0.119pt}{1.638pt}}
\multiput(1222.17,429.00)(13.000,46.599){2}{\rule{0.400pt}{0.819pt}}
\multiput(1236.00,477.93)(0.758,-0.488){13}{\rule{0.700pt}{0.117pt}}
\multiput(1236.00,478.17)(10.547,-8.000){2}{\rule{0.350pt}{0.400pt}}
\multiput(1248.58,471.00)(0.493,1.567){23}{\rule{0.119pt}{1.331pt}}
\multiput(1247.17,471.00)(13.000,37.238){2}{\rule{0.400pt}{0.665pt}}
\multiput(1261.58,511.00)(0.493,0.933){23}{\rule{0.119pt}{0.838pt}}
\multiput(1260.17,511.00)(13.000,22.260){2}{\rule{0.400pt}{0.419pt}}
\multiput(1274.58,535.00)(0.492,1.272){21}{\rule{0.119pt}{1.100pt}}
\multiput(1273.17,535.00)(12.000,27.717){2}{\rule{0.400pt}{0.550pt}}
\multiput(1286.58,565.00)(0.493,1.052){23}{\rule{0.119pt}{0.931pt}}
\multiput(1285.17,565.00)(13.000,25.068){2}{\rule{0.400pt}{0.465pt}}
\multiput(1299.00,592.58)(0.497,0.493){23}{\rule{0.500pt}{0.119pt}}
\multiput(1299.00,591.17)(11.962,13.000){2}{\rule{0.250pt}{0.400pt}}
\multiput(1312.00,605.58)(0.497,0.493){23}{\rule{0.500pt}{0.119pt}}
\multiput(1312.00,604.17)(11.962,13.000){2}{\rule{0.250pt}{0.400pt}}
\multiput(1325.58,618.00)(0.492,1.099){21}{\rule{0.119pt}{0.967pt}}
\multiput(1324.17,618.00)(12.000,23.994){2}{\rule{0.400pt}{0.483pt}}
\multiput(1337.58,641.80)(0.493,-0.536){23}{\rule{0.119pt}{0.531pt}}
\multiput(1336.17,642.90)(13.000,-12.898){2}{\rule{0.400pt}{0.265pt}}
\multiput(1350.00,630.61)(2.695,0.447){3}{\rule{1.833pt}{0.108pt}}
\multiput(1350.00,629.17)(9.195,3.000){2}{\rule{0.917pt}{0.400pt}}
\multiput(1363.58,633.00)(0.492,0.712){21}{\rule{0.119pt}{0.667pt}}
\multiput(1362.17,633.00)(12.000,15.616){2}{\rule{0.400pt}{0.333pt}}
\multiput(1375.58,650.00)(0.493,1.290){23}{\rule{0.119pt}{1.115pt}}
\multiput(1374.17,650.00)(13.000,30.685){2}{\rule{0.400pt}{0.558pt}}
\multiput(1388.58,683.00)(0.493,1.052){23}{\rule{0.119pt}{0.931pt}}
\multiput(1387.17,683.00)(13.000,25.068){2}{\rule{0.400pt}{0.465pt}}
\multiput(1401.58,710.00)(0.493,0.893){23}{\rule{0.119pt}{0.808pt}}
\multiput(1400.17,710.00)(13.000,21.324){2}{\rule{0.400pt}{0.404pt}}
\multiput(1414.58,733.00)(0.492,1.272){21}{\rule{0.119pt}{1.100pt}}
\multiput(1413.17,733.00)(12.000,27.717){2}{\rule{0.400pt}{0.550pt}}
\multiput(1426.58,763.00)(0.493,2.994){23}{\rule{0.119pt}{2.438pt}}
\multiput(1425.17,763.00)(13.000,70.939){2}{\rule{0.400pt}{1.219pt}}
\put(1032.0,435.0){\rule[-0.200pt]{3.132pt}{0.400pt}}
%\put(180.0,82.0){\rule[-0.200pt]{303.293pt}{0.400pt}}
%\put(1439.0,82.0){\rule[-0.200pt]{0.400pt}{187.420pt}}
%\put(180.0,860.0){\rule[-0.200pt]{303.293pt}{0.400pt}}
%\put(180.0,82.0){\rule[-0.200pt]{0.400pt}{187.420pt}}
\end{picture}

%% file: stat9bacc-50.tex
% GNUPLOT: LaTeX picture
\setlength{\unitlength}{0.240900pt}
\setlength{\unitlength}{0.168pt}
\ifx\plotpoint\undefined\newsavebox{\plotpoint}\fi
\sbox{\plotpoint}{\rule[-0.200pt]{0.400pt}{0.400pt}}%
\begin{picture}(1500,900)(0,0)
\sbox{\plotpoint}{\rule[-0.200pt]{0.400pt}{0.400pt}}%
\put(170.0,82.0){\rule[-0.200pt]{4.818pt}{0.400pt}}
\put(150,82){\makebox(0,0)[r]{ 0}}
\put(1430.0,82.0){\rule[-0.200pt]{4.818pt}{0.400pt}}
\put(170.0,238.0){\rule[-0.200pt]{4.818pt}{0.400pt}}
\put(150,238){\makebox(0,0)[r]{ 20}}
\put(1430.0,238.0){\rule[-0.200pt]{4.818pt}{0.400pt}}
\put(170.0,393.0){\rule[-0.200pt]{4.818pt}{0.400pt}}
\put(150,393){\makebox(0,0)[r]{ 40}}
\put(1430.0,393.0){\rule[-0.200pt]{4.818pt}{0.400pt}}
\put(170.0,549.0){\rule[-0.200pt]{4.818pt}{0.400pt}}
\put(150,549){\makebox(0,0)[r]{ 60}}
\put(1430.0,549.0){\rule[-0.200pt]{4.818pt}{0.400pt}}
\put(170.0,704.0){\rule[-0.200pt]{4.818pt}{0.400pt}}
\put(150,704){\makebox(0,0)[r]{ 80}}
\put(1430.0,704.0){\rule[-0.200pt]{4.818pt}{0.400pt}}
\put(170.0,860.0){\rule[-0.200pt]{4.818pt}{0.400pt}}
\put(150,860){\makebox(0,0)[r]{ 100}}
\put(1430.0,860.0){\rule[-0.200pt]{4.818pt}{0.400pt}}
\put(170.0,82.0){\rule[-0.200pt]{0.400pt}{4.818pt}}
\put(170,41){\makebox(0,0){ 0}}
\put(170.0,840.0){\rule[-0.200pt]{0.400pt}{4.818pt}}
\put(299.0,82.0){\rule[-0.200pt]{0.400pt}{4.818pt}}
\put(299,41){\makebox(0,0){ 10}}
\put(299.0,840.0){\rule[-0.200pt]{0.400pt}{4.818pt}}
\put(429.0,82.0){\rule[-0.200pt]{0.400pt}{4.818pt}}
\put(429,41){\makebox(0,0){ 20}}
\put(429.0,840.0){\rule[-0.200pt]{0.400pt}{4.818pt}}
\put(558.0,82.0){\rule[-0.200pt]{0.400pt}{4.818pt}}
\put(558,41){\makebox(0,0){ 30}}
\put(558.0,840.0){\rule[-0.200pt]{0.400pt}{4.818pt}}
\put(687.0,82.0){\rule[-0.200pt]{0.400pt}{4.818pt}}
\put(687,41){\makebox(0,0){ 40}}
\put(687.0,840.0){\rule[-0.200pt]{0.400pt}{4.818pt}}
\put(816.0,82.0){\rule[-0.200pt]{0.400pt}{4.818pt}}
\put(816,41){\makebox(0,0){ 50}}
\put(816.0,840.0){\rule[-0.200pt]{0.400pt}{4.818pt}}
\put(946.0,82.0){\rule[-0.200pt]{0.400pt}{4.818pt}}
\put(946,41){\makebox(0,0){ 60}}
\put(946.0,840.0){\rule[-0.200pt]{0.400pt}{4.818pt}}
\put(1075.0,82.0){\rule[-0.200pt]{0.400pt}{4.818pt}}
\put(1075,41){\makebox(0,0){ 70}}
\put(1075.0,840.0){\rule[-0.200pt]{0.400pt}{4.818pt}}
\put(1204.0,82.0){\rule[-0.200pt]{0.400pt}{4.818pt}}
\put(1204,41){\makebox(0,0){ 80}}
\put(1204.0,840.0){\rule[-0.200pt]{0.400pt}{4.818pt}}
\put(1334.0,82.0){\rule[-0.200pt]{0.400pt}{4.818pt}}
\put(1334,41){\makebox(0,0){ 90}}
\put(1334.0,840.0){\rule[-0.200pt]{0.400pt}{4.818pt}}
%\put(170.0,82.0){\rule[-0.200pt]{0.400pt}{187.420pt}}
\put(170.0,82.0){\rule[-0.200pt]{0.400pt}{132.26pt}}
%\put(170.0,82.0){\rule[-0.200pt]{308.352pt}{0.400pt}}
\put(170.0,82.0){\rule[-0.200pt]{215.04pt}{0.400pt}}
%\put(1450.0,82.0){\rule[-0.200pt]{0.400pt}{187.420pt}}
\put(1450.0,82.0){\rule[-0.200pt]{0.400pt}{132.26pt}}
%\put(170.0,860.0){\rule[-0.200pt]{308.352pt}{0.400pt}}
\put(170.0,860.0){\rule[-0.200pt]{215.04pt}{0.400pt}}
\put(170,860){\usebox{\plotpoint}}
\put(183,858.67){\rule{3.132pt}{0.400pt}}
\multiput(183.00,859.17)(6.500,-1.000){2}{\rule{1.566pt}{0.400pt}}
\put(196,857.17){\rule{2.700pt}{0.400pt}}
\multiput(196.00,858.17)(7.396,-2.000){2}{\rule{1.350pt}{0.400pt}}
\put(209,855.17){\rule{2.700pt}{0.400pt}}
\multiput(209.00,856.17)(7.396,-2.000){2}{\rule{1.350pt}{0.400pt}}
\put(222,853.67){\rule{3.132pt}{0.400pt}}
\multiput(222.00,854.17)(6.500,-1.000){2}{\rule{1.566pt}{0.400pt}}
\put(235,852.67){\rule{3.132pt}{0.400pt}}
\multiput(235.00,853.17)(6.500,-1.000){2}{\rule{1.566pt}{0.400pt}}
\put(248,851.67){\rule{3.132pt}{0.400pt}}
\multiput(248.00,852.17)(6.500,-1.000){2}{\rule{1.566pt}{0.400pt}}
\put(170.0,860.0){\rule[-0.200pt]{3.132pt}{0.400pt}}
\put(312,850.67){\rule{3.132pt}{0.400pt}}
\multiput(312.00,851.17)(6.500,-1.000){2}{\rule{1.566pt}{0.400pt}}
\put(261.0,852.0){\rule[-0.200pt]{12.286pt}{0.400pt}}
\put(338,849.67){\rule{3.132pt}{0.400pt}}
\multiput(338.00,850.17)(6.500,-1.000){2}{\rule{1.566pt}{0.400pt}}
\put(351,848.67){\rule{3.132pt}{0.400pt}}
\multiput(351.00,849.17)(6.500,-1.000){2}{\rule{1.566pt}{0.400pt}}
\put(364,847.67){\rule{3.132pt}{0.400pt}}
\multiput(364.00,848.17)(6.500,-1.000){2}{\rule{1.566pt}{0.400pt}}
\multiput(377.00,846.95)(2.695,-0.447){3}{\rule{1.833pt}{0.108pt}}
\multiput(377.00,847.17)(9.195,-3.000){2}{\rule{0.917pt}{0.400pt}}
\put(390,843.17){\rule{2.700pt}{0.400pt}}
\multiput(390.00,844.17)(7.396,-2.000){2}{\rule{1.350pt}{0.400pt}}
\put(403,841.17){\rule{2.700pt}{0.400pt}}
\multiput(403.00,842.17)(7.396,-2.000){2}{\rule{1.350pt}{0.400pt}}
\multiput(416.00,839.95)(2.695,-0.447){3}{\rule{1.833pt}{0.108pt}}
\multiput(416.00,840.17)(9.195,-3.000){2}{\rule{0.917pt}{0.400pt}}
\multiput(429.00,836.95)(2.695,-0.447){3}{\rule{1.833pt}{0.108pt}}
\multiput(429.00,837.17)(9.195,-3.000){2}{\rule{0.917pt}{0.400pt}}
\multiput(442.00,833.93)(1.267,-0.477){7}{\rule{1.060pt}{0.115pt}}
\multiput(442.00,834.17)(9.800,-5.000){2}{\rule{0.530pt}{0.400pt}}
\multiput(454.00,828.94)(1.797,-0.468){5}{\rule{1.400pt}{0.113pt}}
\multiput(454.00,829.17)(10.094,-4.000){2}{\rule{0.700pt}{0.400pt}}
\multiput(467.00,824.93)(1.378,-0.477){7}{\rule{1.140pt}{0.115pt}}
\multiput(467.00,825.17)(10.634,-5.000){2}{\rule{0.570pt}{0.400pt}}
\multiput(480.00,819.95)(2.695,-0.447){3}{\rule{1.833pt}{0.108pt}}
\multiput(480.00,820.17)(9.195,-3.000){2}{\rule{0.917pt}{0.400pt}}
\multiput(493.00,816.94)(1.797,-0.468){5}{\rule{1.400pt}{0.113pt}}
\multiput(493.00,817.17)(10.094,-4.000){2}{\rule{0.700pt}{0.400pt}}
\put(506,812.17){\rule{2.700pt}{0.400pt}}
\multiput(506.00,813.17)(7.396,-2.000){2}{\rule{1.350pt}{0.400pt}}
\put(519,810.17){\rule{2.700pt}{0.400pt}}
\multiput(519.00,811.17)(7.396,-2.000){2}{\rule{1.350pt}{0.400pt}}
\put(532,808.67){\rule{3.132pt}{0.400pt}}
\multiput(532.00,809.17)(6.500,-1.000){2}{\rule{1.566pt}{0.400pt}}
\put(545,807.67){\rule{3.132pt}{0.400pt}}
\multiput(545.00,808.17)(6.500,-1.000){2}{\rule{1.566pt}{0.400pt}}
\put(558,806.67){\rule{3.132pt}{0.400pt}}
\multiput(558.00,807.17)(6.500,-1.000){2}{\rule{1.566pt}{0.400pt}}
\put(571,805.67){\rule{3.132pt}{0.400pt}}
\multiput(571.00,806.17)(6.500,-1.000){2}{\rule{1.566pt}{0.400pt}}
\put(584,804.67){\rule{3.132pt}{0.400pt}}
\multiput(584.00,805.17)(6.500,-1.000){2}{\rule{1.566pt}{0.400pt}}
\put(597,803.67){\rule{3.132pt}{0.400pt}}
\multiput(597.00,804.17)(6.500,-1.000){2}{\rule{1.566pt}{0.400pt}}
\put(610,802.67){\rule{3.132pt}{0.400pt}}
\multiput(610.00,803.17)(6.500,-1.000){2}{\rule{1.566pt}{0.400pt}}
\put(623,801.67){\rule{2.891pt}{0.400pt}}
\multiput(623.00,802.17)(6.000,-1.000){2}{\rule{1.445pt}{0.400pt}}
\put(635,800.67){\rule{3.132pt}{0.400pt}}
\multiput(635.00,801.17)(6.500,-1.000){2}{\rule{1.566pt}{0.400pt}}
\put(648,799.17){\rule{2.700pt}{0.400pt}}
\multiput(648.00,800.17)(7.396,-2.000){2}{\rule{1.350pt}{0.400pt}}
\put(661,797.17){\rule{2.700pt}{0.400pt}}
\multiput(661.00,798.17)(7.396,-2.000){2}{\rule{1.350pt}{0.400pt}}
\put(674,795.17){\rule{2.700pt}{0.400pt}}
\multiput(674.00,796.17)(7.396,-2.000){2}{\rule{1.350pt}{0.400pt}}
\multiput(687.00,793.95)(2.695,-0.447){3}{\rule{1.833pt}{0.108pt}}
\multiput(687.00,794.17)(9.195,-3.000){2}{\rule{0.917pt}{0.400pt}}
\multiput(700.00,790.95)(2.695,-0.447){3}{\rule{1.833pt}{0.108pt}}
\multiput(700.00,791.17)(9.195,-3.000){2}{\rule{0.917pt}{0.400pt}}
\multiput(713.00,787.95)(2.695,-0.447){3}{\rule{1.833pt}{0.108pt}}
\multiput(713.00,788.17)(9.195,-3.000){2}{\rule{0.917pt}{0.400pt}}
\multiput(726.00,784.95)(2.695,-0.447){3}{\rule{1.833pt}{0.108pt}}
\multiput(726.00,785.17)(9.195,-3.000){2}{\rule{0.917pt}{0.400pt}}
\multiput(739.00,781.94)(1.797,-0.468){5}{\rule{1.400pt}{0.113pt}}
\multiput(739.00,782.17)(10.094,-4.000){2}{\rule{0.700pt}{0.400pt}}
\multiput(752.00,777.94)(1.797,-0.468){5}{\rule{1.400pt}{0.113pt}}
\multiput(752.00,778.17)(10.094,-4.000){2}{\rule{0.700pt}{0.400pt}}
\multiput(765.00,773.93)(1.378,-0.477){7}{\rule{1.140pt}{0.115pt}}
\multiput(765.00,774.17)(10.634,-5.000){2}{\rule{0.570pt}{0.400pt}}
\multiput(778.00,768.93)(1.378,-0.477){7}{\rule{1.140pt}{0.115pt}}
\multiput(778.00,769.17)(10.634,-5.000){2}{\rule{0.570pt}{0.400pt}}
\multiput(791.00,763.93)(1.123,-0.482){9}{\rule{0.967pt}{0.116pt}}
\multiput(791.00,764.17)(10.994,-6.000){2}{\rule{0.483pt}{0.400pt}}
\multiput(804.00,757.93)(0.874,-0.485){11}{\rule{0.786pt}{0.117pt}}
\multiput(804.00,758.17)(10.369,-7.000){2}{\rule{0.393pt}{0.400pt}}
\multiput(816.00,750.93)(0.728,-0.489){15}{\rule{0.678pt}{0.118pt}}
\multiput(816.00,751.17)(11.593,-9.000){2}{\rule{0.339pt}{0.400pt}}
\multiput(829.00,741.93)(0.728,-0.489){15}{\rule{0.678pt}{0.118pt}}
\multiput(829.00,742.17)(11.593,-9.000){2}{\rule{0.339pt}{0.400pt}}
\multiput(842.00,732.93)(0.824,-0.488){13}{\rule{0.750pt}{0.117pt}}
\multiput(842.00,733.17)(11.443,-8.000){2}{\rule{0.375pt}{0.400pt}}
\multiput(855.00,724.92)(0.652,-0.491){17}{\rule{0.620pt}{0.118pt}}
\multiput(855.00,725.17)(11.713,-10.000){2}{\rule{0.310pt}{0.400pt}}
\multiput(868.00,714.93)(0.824,-0.488){13}{\rule{0.750pt}{0.117pt}}
\multiput(868.00,715.17)(11.443,-8.000){2}{\rule{0.375pt}{0.400pt}}
\multiput(881.00,706.93)(0.728,-0.489){15}{\rule{0.678pt}{0.118pt}}
\multiput(881.00,707.17)(11.593,-9.000){2}{\rule{0.339pt}{0.400pt}}
\multiput(894.00,697.93)(0.824,-0.488){13}{\rule{0.750pt}{0.117pt}}
\multiput(894.00,698.17)(11.443,-8.000){2}{\rule{0.375pt}{0.400pt}}
\multiput(907.00,689.93)(0.728,-0.489){15}{\rule{0.678pt}{0.118pt}}
\multiput(907.00,690.17)(11.593,-9.000){2}{\rule{0.339pt}{0.400pt}}
\multiput(920.00,680.92)(0.652,-0.491){17}{\rule{0.620pt}{0.118pt}}
\multiput(920.00,681.17)(11.713,-10.000){2}{\rule{0.310pt}{0.400pt}}
\multiput(933.00,670.93)(0.824,-0.488){13}{\rule{0.750pt}{0.117pt}}
\multiput(933.00,671.17)(11.443,-8.000){2}{\rule{0.375pt}{0.400pt}}
\multiput(946.00,662.93)(0.950,-0.485){11}{\rule{0.843pt}{0.117pt}}
\multiput(946.00,663.17)(11.251,-7.000){2}{\rule{0.421pt}{0.400pt}}
\multiput(959.00,655.93)(0.950,-0.485){11}{\rule{0.843pt}{0.117pt}}
\multiput(959.00,656.17)(11.251,-7.000){2}{\rule{0.421pt}{0.400pt}}
\multiput(972.00,648.93)(1.123,-0.482){9}{\rule{0.967pt}{0.116pt}}
\multiput(972.00,649.17)(10.994,-6.000){2}{\rule{0.483pt}{0.400pt}}
\multiput(985.00,642.93)(1.033,-0.482){9}{\rule{0.900pt}{0.116pt}}
\multiput(985.00,643.17)(10.132,-6.000){2}{\rule{0.450pt}{0.400pt}}
\multiput(997.00,636.93)(1.378,-0.477){7}{\rule{1.140pt}{0.115pt}}
\multiput(997.00,637.17)(10.634,-5.000){2}{\rule{0.570pt}{0.400pt}}
\multiput(1010.00,631.93)(1.378,-0.477){7}{\rule{1.140pt}{0.115pt}}
\multiput(1010.00,632.17)(10.634,-5.000){2}{\rule{0.570pt}{0.400pt}}
\multiput(1023.00,626.93)(1.123,-0.482){9}{\rule{0.967pt}{0.116pt}}
\multiput(1023.00,627.17)(10.994,-6.000){2}{\rule{0.483pt}{0.400pt}}
\multiput(1036.00,620.95)(2.695,-0.447){3}{\rule{1.833pt}{0.108pt}}
\multiput(1036.00,621.17)(9.195,-3.000){2}{\rule{0.917pt}{0.400pt}}
\multiput(1049.00,617.94)(1.797,-0.468){5}{\rule{1.400pt}{0.113pt}}
\multiput(1049.00,618.17)(10.094,-4.000){2}{\rule{0.700pt}{0.400pt}}
\multiput(1062.00,613.95)(2.695,-0.447){3}{\rule{1.833pt}{0.108pt}}
\multiput(1062.00,614.17)(9.195,-3.000){2}{\rule{0.917pt}{0.400pt}}
\multiput(1075.00,610.95)(2.695,-0.447){3}{\rule{1.833pt}{0.108pt}}
\multiput(1075.00,611.17)(9.195,-3.000){2}{\rule{0.917pt}{0.400pt}}
\multiput(1088.00,607.95)(2.695,-0.447){3}{\rule{1.833pt}{0.108pt}}
\multiput(1088.00,608.17)(9.195,-3.000){2}{\rule{0.917pt}{0.400pt}}
\put(1101,604.17){\rule{2.700pt}{0.400pt}}
\multiput(1101.00,605.17)(7.396,-2.000){2}{\rule{1.350pt}{0.400pt}}
\multiput(1114.00,602.95)(2.695,-0.447){3}{\rule{1.833pt}{0.108pt}}
\multiput(1114.00,603.17)(9.195,-3.000){2}{\rule{0.917pt}{0.400pt}}
\put(1127,599.17){\rule{2.700pt}{0.400pt}}
\multiput(1127.00,600.17)(7.396,-2.000){2}{\rule{1.350pt}{0.400pt}}
\put(1140,597.17){\rule{2.700pt}{0.400pt}}
\multiput(1140.00,598.17)(7.396,-2.000){2}{\rule{1.350pt}{0.400pt}}
\multiput(1153.00,595.95)(2.695,-0.447){3}{\rule{1.833pt}{0.108pt}}
\multiput(1153.00,596.17)(9.195,-3.000){2}{\rule{0.917pt}{0.400pt}}
\multiput(1166.00,592.95)(2.472,-0.447){3}{\rule{1.700pt}{0.108pt}}
\multiput(1166.00,593.17)(8.472,-3.000){2}{\rule{0.850pt}{0.400pt}}
\put(1178,589.17){\rule{2.700pt}{0.400pt}}
\multiput(1178.00,590.17)(7.396,-2.000){2}{\rule{1.350pt}{0.400pt}}
\multiput(1191.00,587.94)(1.797,-0.468){5}{\rule{1.400pt}{0.113pt}}
\multiput(1191.00,588.17)(10.094,-4.000){2}{\rule{0.700pt}{0.400pt}}
\multiput(1204.00,583.95)(2.695,-0.447){3}{\rule{1.833pt}{0.108pt}}
\multiput(1204.00,584.17)(9.195,-3.000){2}{\rule{0.917pt}{0.400pt}}
\multiput(1217.00,580.95)(2.695,-0.447){3}{\rule{1.833pt}{0.108pt}}
\multiput(1217.00,581.17)(9.195,-3.000){2}{\rule{0.917pt}{0.400pt}}
\multiput(1230.00,577.94)(1.797,-0.468){5}{\rule{1.400pt}{0.113pt}}
\multiput(1230.00,578.17)(10.094,-4.000){2}{\rule{0.700pt}{0.400pt}}
\multiput(1243.00,573.93)(1.378,-0.477){7}{\rule{1.140pt}{0.115pt}}
\multiput(1243.00,574.17)(10.634,-5.000){2}{\rule{0.570pt}{0.400pt}}
\multiput(1256.00,568.93)(1.378,-0.477){7}{\rule{1.140pt}{0.115pt}}
\multiput(1256.00,569.17)(10.634,-5.000){2}{\rule{0.570pt}{0.400pt}}
\multiput(1269.00,563.93)(0.950,-0.485){11}{\rule{0.843pt}{0.117pt}}
\multiput(1269.00,564.17)(11.251,-7.000){2}{\rule{0.421pt}{0.400pt}}
\multiput(1282.00,556.93)(0.728,-0.489){15}{\rule{0.678pt}{0.118pt}}
\multiput(1282.00,557.17)(11.593,-9.000){2}{\rule{0.339pt}{0.400pt}}
\multiput(1295.00,547.92)(0.590,-0.492){19}{\rule{0.573pt}{0.118pt}}
\multiput(1295.00,548.17)(11.811,-11.000){2}{\rule{0.286pt}{0.400pt}}
\multiput(1308.58,535.80)(0.493,-0.536){23}{\rule{0.119pt}{0.531pt}}
\multiput(1307.17,536.90)(13.000,-12.898){2}{\rule{0.400pt}{0.265pt}}
\multiput(1321.58,521.54)(0.493,-0.616){23}{\rule{0.119pt}{0.592pt}}
\multiput(1320.17,522.77)(13.000,-14.771){2}{\rule{0.400pt}{0.296pt}}
\multiput(1334.58,505.16)(0.493,-0.734){23}{\rule{0.119pt}{0.685pt}}
\multiput(1333.17,506.58)(13.000,-17.579){2}{\rule{0.400pt}{0.342pt}}
\multiput(1347.58,485.40)(0.492,-0.970){21}{\rule{0.119pt}{0.867pt}}
\multiput(1346.17,487.20)(12.000,-21.201){2}{\rule{0.400pt}{0.433pt}}
\multiput(1359.58,463.03)(0.493,-0.774){23}{\rule{0.119pt}{0.715pt}}
\multiput(1358.17,464.52)(13.000,-18.515){2}{\rule{0.400pt}{0.358pt}}
\multiput(1372.58,443.03)(0.493,-0.774){23}{\rule{0.119pt}{0.715pt}}
\multiput(1371.17,444.52)(13.000,-18.515){2}{\rule{0.400pt}{0.358pt}}
\multiput(1385.58,422.52)(0.493,-0.933){23}{\rule{0.119pt}{0.838pt}}
\multiput(1384.17,424.26)(13.000,-22.260){2}{\rule{0.400pt}{0.419pt}}
\multiput(1398.58,397.37)(0.493,-1.290){23}{\rule{0.119pt}{1.115pt}}
\multiput(1397.17,399.68)(13.000,-30.685){2}{\rule{0.400pt}{0.558pt}}
\multiput(1411.58,363.35)(0.493,-1.607){23}{\rule{0.119pt}{1.362pt}}
\multiput(1410.17,366.17)(13.000,-38.174){2}{\rule{0.400pt}{0.681pt}}
\multiput(1424.58,321.20)(0.493,-1.964){23}{\rule{0.119pt}{1.638pt}}
\multiput(1423.17,324.60)(13.000,-46.599){2}{\rule{0.400pt}{0.819pt}}
\multiput(1437.58,269.15)(0.493,-2.598){23}{\rule{0.119pt}{2.131pt}}
\multiput(1436.17,273.58)(13.000,-61.577){2}{\rule{0.400pt}{1.065pt}}
\put(325.0,851.0){\rule[-0.200pt]{3.132pt}{0.400pt}}
%\put(170.0,82.0){\rule[-0.200pt]{0.400pt}{187.420pt}}
%\put(170.0,82.0){\rule[-0.200pt]{308.352pt}{0.400pt}}
%\put(1450.0,82.0){\rule[-0.200pt]{0.400pt}{187.420pt}}
%\put(170.0,860.0){\rule[-0.200pt]{308.352pt}{0.400pt}}
\end{picture}

%% file: stat9e-rnk.tex
% GNUPLOT: LaTeX picture
%\setlength{\unitlength}{0.240900pt}
 \setlength{\unitlength}{0.168pt}
\ifx\plotpoint\undefined\newsavebox{\plotpoint}\fi
\sbox{\plotpoint}{\rule[-0.200pt]{0.400pt}{0.400pt}}%
\begin{picture}(1500,900)(0,0)
\sbox{\plotpoint}{\rule[-0.200pt]{0.400pt}{0.400pt}}%
\put(140.0,82.0){\rule[-0.200pt]{4.818pt}{0.400pt}}
\put(120,82){\makebox(0,0)[r]{ 0}}
\put(1419.0,82.0){\rule[-0.200pt]{4.818pt}{0.400pt}}
\put(140.0,160.0){\rule[-0.200pt]{4.818pt}{0.400pt}}
\put(120,160){\makebox(0,0)[r]{ 10}}
\put(1419.0,160.0){\rule[-0.200pt]{4.818pt}{0.400pt}}
\put(140.0,238.0){\rule[-0.200pt]{4.818pt}{0.400pt}}
\put(120,238){\makebox(0,0)[r]{ 20}}
\put(1419.0,238.0){\rule[-0.200pt]{4.818pt}{0.400pt}}
\put(140.0,315.0){\rule[-0.200pt]{4.818pt}{0.400pt}}
\put(120,315){\makebox(0,0)[r]{ 30}}
\put(1419.0,315.0){\rule[-0.200pt]{4.818pt}{0.400pt}}
\put(140.0,393.0){\rule[-0.200pt]{4.818pt}{0.400pt}}
\put(120,393){\makebox(0,0)[r]{ 40}}
\put(1419.0,393.0){\rule[-0.200pt]{4.818pt}{0.400pt}}
\put(140.0,471.0){\rule[-0.200pt]{4.818pt}{0.400pt}}
\put(120,471){\makebox(0,0)[r]{ 50}}
\put(1419.0,471.0){\rule[-0.200pt]{4.818pt}{0.400pt}}
\put(140.0,549.0){\rule[-0.200pt]{4.818pt}{0.400pt}}
\put(120,549){\makebox(0,0)[r]{ 60}}
\put(1419.0,549.0){\rule[-0.200pt]{4.818pt}{0.400pt}}
\put(140.0,627.0){\rule[-0.200pt]{4.818pt}{0.400pt}}
\put(120,627){\makebox(0,0)[r]{ 70}}
\put(1419.0,627.0){\rule[-0.200pt]{4.818pt}{0.400pt}}
\put(140.0,704.0){\rule[-0.200pt]{4.818pt}{0.400pt}}
\put(120,704){\makebox(0,0)[r]{ 80}}
\put(1419.0,704.0){\rule[-0.200pt]{4.818pt}{0.400pt}}
\put(140.0,782.0){\rule[-0.200pt]{4.818pt}{0.400pt}}
\put(120,782){\makebox(0,0)[r]{ 90}}
\put(1419.0,782.0){\rule[-0.200pt]{4.818pt}{0.400pt}}
\put(140.0,860.0){\rule[-0.200pt]{4.818pt}{0.400pt}}
\put(120,860){\makebox(0,0)[r]{ 100}}
\put(1419.0,860.0){\rule[-0.200pt]{4.818pt}{0.400pt}}
\put(140.0,82.0){\rule[-0.200pt]{0.400pt}{4.818pt}}
\put(140,41){\makebox(0,0){ 0}}
\put(140.0,840.0){\rule[-0.200pt]{0.400pt}{4.818pt}}
\put(356.0,82.0){\rule[-0.200pt]{0.400pt}{4.818pt}}
\put(356,41){\makebox(0,0){ 5}}
\put(356.0,840.0){\rule[-0.200pt]{0.400pt}{4.818pt}}
\put(573.0,82.0){\rule[-0.200pt]{0.400pt}{4.818pt}}
\put(573,41){\makebox(0,0){ 10}}
\put(573.0,840.0){\rule[-0.200pt]{0.400pt}{4.818pt}}
\put(789.0,82.0){\rule[-0.200pt]{0.400pt}{4.818pt}}
\put(789,41){\makebox(0,0){ 15}}
\put(789.0,840.0){\rule[-0.200pt]{0.400pt}{4.818pt}}
\put(1006.0,82.0){\rule[-0.200pt]{0.400pt}{4.818pt}}
\put(1006,41){\makebox(0,0){ 20}}
\put(1006.0,840.0){\rule[-0.200pt]{0.400pt}{4.818pt}}
\put(1222.0,82.0){\rule[-0.200pt]{0.400pt}{4.818pt}}
\put(1222,41){\makebox(0,0){ 25}}
\put(1222.0,840.0){\rule[-0.200pt]{0.400pt}{4.818pt}}
\put(1439.0,82.0){\rule[-0.200pt]{0.400pt}{4.818pt}}
\put(1439,41){\makebox(0,0){ 30}}
\put(1439.0,840.0){\rule[-0.200pt]{0.400pt}{4.818pt}}
%\put(140.0,82.0){\rule[-0.200pt]{0.400pt}{187.420pt}}
%\put(140.0,82.0){\rule[-0.200pt]{312.929pt}{0.400pt}}
%\put(1439.0,82.0){\rule[-0.200pt]{0.400pt}{187.420pt}}
%\put(140.0,860.0){\rule[-0.200pt]{312.929pt}{0.400pt}}
 \put(140.0,82.0){\rule[-0.200pt]{0.400pt}{132.260pt}}
 \put(140.0,82.0){\rule[-0.200pt]{217.430pt}{0.400pt}}
 \put(1439.0,82.0){\rule[-0.200pt]{0.400pt}{132.260pt}}
 \put(140.0,860.0){\rule[-0.200pt]{217.430pt}{0.400pt}}
\put(1279,320){\makebox(0,0)[r]{\scriptsize SEDS}}
\put(1299.0,320.0){\rule[-0.200pt]{18pt}{0.400pt}}
\put(183,131){\usebox{\plotpoint}}
\multiput(183.00,131.58)(0.647,0.498){65}{\rule{0.618pt}{0.120pt}}
\multiput(183.00,130.17)(42.718,34.000){2}{\rule{0.309pt}{0.400pt}}
\multiput(227.00,165.58)(0.770,0.497){53}{\rule{0.714pt}{0.120pt}}
\multiput(227.00,164.17)(41.517,28.000){2}{\rule{0.357pt}{0.400pt}}
\multiput(270.00,193.58)(0.799,0.497){51}{\rule{0.737pt}{0.120pt}}
\multiput(270.00,192.17)(41.470,27.000){2}{\rule{0.369pt}{0.400pt}}
\multiput(313.00,220.58)(1.031,0.496){39}{\rule{0.919pt}{0.119pt}}
\multiput(313.00,219.17)(41.092,21.000){2}{\rule{0.460pt}{0.400pt}}
\multiput(356.00,241.58)(1.055,0.496){39}{\rule{0.938pt}{0.119pt}}
\multiput(356.00,240.17)(42.053,21.000){2}{\rule{0.469pt}{0.400pt}}
\multiput(400.00,262.58)(1.207,0.495){33}{\rule{1.056pt}{0.119pt}}
\multiput(400.00,261.17)(40.809,18.000){2}{\rule{0.528pt}{0.400pt}}
\multiput(443.00,280.58)(1.562,0.494){25}{\rule{1.329pt}{0.119pt}}
\multiput(443.00,279.17)(40.242,14.000){2}{\rule{0.664pt}{0.400pt}}
\multiput(486.00,294.58)(1.726,0.493){23}{\rule{1.454pt}{0.119pt}}
\multiput(486.00,293.17)(40.982,13.000){2}{\rule{0.727pt}{0.400pt}}
\multiput(530.00,307.58)(1.686,0.493){23}{\rule{1.423pt}{0.119pt}}
\multiput(530.00,306.17)(40.046,13.000){2}{\rule{0.712pt}{0.400pt}}
\multiput(573.00,320.58)(1.455,0.494){27}{\rule{1.247pt}{0.119pt}}
\multiput(573.00,319.17)(40.412,15.000){2}{\rule{0.623pt}{0.400pt}}
\multiput(616.00,335.58)(1.875,0.492){21}{\rule{1.567pt}{0.119pt}}
\multiput(616.00,334.17)(40.748,12.000){2}{\rule{0.783pt}{0.400pt}}
\multiput(660.00,347.58)(1.686,0.493){23}{\rule{1.423pt}{0.119pt}}
\multiput(660.00,346.17)(40.046,13.000){2}{\rule{0.712pt}{0.400pt}}
\multiput(703.00,360.58)(2.215,0.491){17}{\rule{1.820pt}{0.118pt}}
\multiput(703.00,359.17)(39.222,10.000){2}{\rule{0.910pt}{0.400pt}}
\multiput(746.00,370.59)(2.805,0.488){13}{\rule{2.250pt}{0.117pt}}
\multiput(746.00,369.17)(38.330,8.000){2}{\rule{1.125pt}{0.400pt}}
\multiput(789.00,378.58)(2.052,0.492){19}{\rule{1.700pt}{0.118pt}}
\multiput(789.00,377.17)(40.472,11.000){2}{\rule{0.850pt}{0.400pt}}
\multiput(833.00,389.58)(2.215,0.491){17}{\rule{1.820pt}{0.118pt}}
\multiput(833.00,388.17)(39.222,10.000){2}{\rule{0.910pt}{0.400pt}}
\multiput(876.00,399.58)(2.215,0.491){17}{\rule{1.820pt}{0.118pt}}
\multiput(876.00,398.17)(39.222,10.000){2}{\rule{0.910pt}{0.400pt}}
\multiput(919.00,409.59)(2.533,0.489){15}{\rule{2.056pt}{0.118pt}}
\multiput(919.00,408.17)(39.734,9.000){2}{\rule{1.028pt}{0.400pt}}
\multiput(963.00,418.59)(2.805,0.488){13}{\rule{2.250pt}{0.117pt}}
\multiput(963.00,417.17)(38.330,8.000){2}{\rule{1.125pt}{0.400pt}}
\multiput(1006.00,426.59)(2.475,0.489){15}{\rule{2.011pt}{0.118pt}}
\multiput(1006.00,425.17)(38.826,9.000){2}{\rule{1.006pt}{0.400pt}}
\multiput(1049.00,435.59)(4.829,0.477){7}{\rule{3.620pt}{0.115pt}}
\multiput(1049.00,434.17)(36.487,5.000){2}{\rule{1.810pt}{0.400pt}}
\multiput(1093.00,440.59)(3.836,0.482){9}{\rule{2.967pt}{0.116pt}}
\multiput(1093.00,439.17)(36.843,6.000){2}{\rule{1.483pt}{0.400pt}}
\multiput(1136.00,446.59)(3.836,0.482){9}{\rule{2.967pt}{0.116pt}}
\multiput(1136.00,445.17)(36.843,6.000){2}{\rule{1.483pt}{0.400pt}}
\multiput(1179.00,452.59)(3.836,0.482){9}{\rule{2.967pt}{0.116pt}}
\multiput(1179.00,451.17)(36.843,6.000){2}{\rule{1.483pt}{0.400pt}}
\multiput(1222.00,458.59)(2.533,0.489){15}{\rule{2.056pt}{0.118pt}}
\multiput(1222.00,457.17)(39.734,9.000){2}{\rule{1.028pt}{0.400pt}}
\multiput(1266.00,467.59)(3.836,0.482){9}{\rule{2.967pt}{0.116pt}}
\multiput(1266.00,466.17)(36.843,6.000){2}{\rule{1.483pt}{0.400pt}}
\multiput(1309.00,473.58)(2.215,0.491){17}{\rule{1.820pt}{0.118pt}}
\multiput(1309.00,472.17)(39.222,10.000){2}{\rule{0.910pt}{0.400pt}}
\multiput(1352.00,483.59)(3.315,0.485){11}{\rule{2.614pt}{0.117pt}}
\multiput(1352.00,482.17)(38.574,7.000){2}{\rule{1.307pt}{0.400pt}}
\multiput(1396.00,490.59)(4.718,0.477){7}{\rule{3.540pt}{0.115pt}}
\multiput(1396.00,489.17)(35.653,5.000){2}{\rule{1.770pt}{0.400pt}}
\put(1439,495){\usebox{\plotpoint}}
\put(1279,279){\makebox(0,0)[r]{\scriptsize EvdW2010}}
\multiput(1299,279)(20,0.000){5}{\usebox{\plotpoint}}
\put(1399,279){\usebox{\plotpoint}}
\put(183,377){\usebox{\plotpoint}}
\multiput(183,377)(8.574,18.902){6}{\usebox{\plotpoint}}
\multiput(227,474)(12.500,16.569){3}{\usebox{\plotpoint}}
\multiput(270,531)(15.733,13.538){3}{\usebox{\plotpoint}}
\multiput(313,568)(17.393,11.326){2}{\usebox{\plotpoint}}
\multiput(356,596)(18.731,8.940){3}{\usebox{\plotpoint}}
\multiput(400,617)(18.819,8.753){2}{\usebox{\plotpoint}}
\multiput(443,637)(19.453,7.238){2}{\usebox{\plotpoint}}
\multiput(486,653)(19.778,6.293){2}{\usebox{\plotpoint}}
\multiput(530,667)(20.108,5.144){2}{\usebox{\plotpoint}}
\multiput(573,678)(20.108,5.144){3}{\usebox{\plotpoint}}
\multiput(616,689)(20.334,4.159){2}{\usebox{\plotpoint}}
\multiput(660,698)(20.405,3.796){2}{\usebox{\plotpoint}}
\multiput(703,706)(20.405,3.796){2}{\usebox{\plotpoint}}
\multiput(746,714)(20.486,3.335){2}{\usebox{\plotpoint}}
\multiput(789,721)(20.498,3.261){2}{\usebox{\plotpoint}}
\multiput(833,728)(20.556,2.868){2}{\usebox{\plotpoint}}
\multiput(876,734)(20.556,2.868){2}{\usebox{\plotpoint}}
\multiput(919,740)(20.623,2.343){3}{\usebox{\plotpoint}}
\multiput(963,745)(20.617,2.397){2}{\usebox{\plotpoint}}
\multiput(1006,750)(20.617,2.397){2}{\usebox{\plotpoint}}
\multiput(1049,755)(20.670,1.879){2}{\usebox{\plotpoint}}
\multiput(1093,759)(20.666,1.922){2}{\usebox{\plotpoint}}
\multiput(1136,763)(20.617,2.397){2}{\usebox{\plotpoint}}
\multiput(1179,768)(20.666,1.922){2}{\usebox{\plotpoint}}
\multiput(1222,772)(20.670,1.879){2}{\usebox{\plotpoint}}
\multiput(1266,776)(20.705,1.445){2}{\usebox{\plotpoint}}
\multiput(1309,779)(20.705,1.445){2}{\usebox{\plotpoint}}
\multiput(1352,782)(20.670,1.879){3}{\usebox{\plotpoint}}
\multiput(1396,786)(20.705,1.445){2}{\usebox{\plotpoint}}
\put(1439,789){\usebox{\plotpoint}}
\sbox{\plotpoint}{\rule[-0.400pt]{0.800pt}{0.800pt}}%
\sbox{\plotpoint}{\rule[-0.200pt]{0.400pt}{0.400pt}}%
\put(1279,238){\makebox(0,0)[r]{\scriptsize SuNa}}
\sbox{\plotpoint}{\rule[-0.400pt]{0.800pt}{0.800pt}}%
\put(1299.0,238.0){\rule[-0.400pt]{18pt}{0.800pt}}
\put(183,368){\usebox{\plotpoint}}
\multiput(184.41,368.00)(0.502,0.996){97}{\rule{0.121pt}{1.785pt}}
\multiput(181.34,368.00)(52.000,99.296){2}{\rule{0.800pt}{0.892pt}}
\multiput(236.41,471.00)(0.503,0.614){63}{\rule{0.121pt}{1.183pt}}
\multiput(233.34,471.00)(35.000,40.545){2}{\rule{0.800pt}{0.591pt}}
\multiput(270.00,515.41)(0.614,0.503){63}{\rule{1.183pt}{0.121pt}}
\multiput(270.00,512.34)(40.545,35.000){2}{\rule{0.591pt}{0.800pt}}
\multiput(313.00,550.41)(1.155,0.506){31}{\rule{2.011pt}{0.122pt}}
\multiput(313.00,547.34)(38.827,19.000){2}{\rule{1.005pt}{0.800pt}}
\multiput(356.00,569.41)(1.413,0.503){71}{\rule{2.436pt}{0.121pt}}
\multiput(356.00,566.34)(103.944,39.000){2}{\rule{1.218pt}{0.800pt}}
\multiput(465.00,608.41)(1.668,0.505){33}{\rule{2.800pt}{0.122pt}}
\multiput(465.00,605.34)(59.188,20.000){2}{\rule{1.400pt}{0.800pt}}
\multiput(530.00,628.40)(3.628,0.526){7}{\rule{5.114pt}{0.127pt}}
\multiput(530.00,625.34)(32.385,7.000){2}{\rule{2.557pt}{0.800pt}}
\multiput(573.00,635.41)(2.139,0.502){95}{\rule{3.588pt}{0.121pt}}
\multiput(573.00,632.34)(208.552,51.000){2}{\rule{1.794pt}{0.800pt}}
\multiput(789.00,686.41)(3.155,0.503){63}{\rule{5.160pt}{0.121pt}}
\multiput(789.00,683.34)(206.290,35.000){2}{\rule{2.580pt}{0.800pt}}
\multiput(1006.00,721.41)(4.263,0.504){45}{\rule{6.846pt}{0.121pt}}
\multiput(1006.00,718.34)(201.790,26.000){2}{\rule{3.423pt}{0.800pt}}
\multiput(1222.00,747.41)(5.349,0.505){35}{\rule{8.467pt}{0.122pt}}
\multiput(1222.00,744.34)(199.427,21.000){2}{\rule{4.233pt}{0.800pt}}
\put(1439,767){\usebox{\plotpoint}}
\sbox{\plotpoint}{\rule[-0.500pt]{1.000pt}{1.000pt}}%
\sbox{\plotpoint}{\rule[-0.200pt]{0.400pt}{0.400pt}}%
\put(1279,197){\makebox(0,0)[r]{\scriptsize SHG}}
\sbox{\plotpoint}{\rule[-0.500pt]{1.000pt}{1.000pt}}%
\multiput(1299,197)(20,0.000){5}{\usebox{\plotpoint}}
\put(1399,197){\usebox{\plotpoint}}
\put(183,373){\usebox{\plotpoint}}
\multiput(183,373)(8.574,18.902){6}{\usebox{\plotpoint}}
\multiput(227,470)(12.929,16.237){3}{\usebox{\plotpoint}}
\multiput(270,524)(15.914,13.324){3}{\usebox{\plotpoint}}
\multiput(313,560)(17.393,11.326){2}{\usebox{\plotpoint}}
\multiput(356,588)(18.564,9.282){3}{\usebox{\plotpoint}}
\multiput(400,610)(19.146,8.014){2}{\usebox{\plotpoint}}
\multiput(443,628)(19.597,6.836){2}{\usebox{\plotpoint}}
\multiput(486,643)(19.778,6.293){2}{\usebox{\plotpoint}}
\multiput(530,657)(20.108,5.144){2}{\usebox{\plotpoint}}
\multiput(573,668)(20.108,5.144){2}{\usebox{\plotpoint}}
\multiput(616,679)(20.239,4.600){3}{\usebox{\plotpoint}}
\multiput(660,689)(20.315,4.252){2}{\usebox{\plotpoint}}
\multiput(703,698)(20.405,3.796){2}{\usebox{\plotpoint}}
\multiput(746,706)(20.405,3.796){2}{\usebox{\plotpoint}}
\multiput(789,714)(20.498,3.261){2}{\usebox{\plotpoint}}
\multiput(833,721)(20.486,3.335){2}{\usebox{\plotpoint}}
\multiput(876,728)(20.556,2.868){2}{\usebox{\plotpoint}}
\multiput(919,734)(20.565,2.804){2}{\usebox{\plotpoint}}
\multiput(963,740)(20.617,2.397){3}{\usebox{\plotpoint}}
\multiput(1006,745)(20.617,2.397){2}{\usebox{\plotpoint}}
\multiput(1049,750)(20.623,2.343){2}{\usebox{\plotpoint}}
\multiput(1093,755)(20.617,2.397){2}{\usebox{\plotpoint}}
\multiput(1136,760)(20.666,1.922){2}{\usebox{\plotpoint}}
\multiput(1179,764)(20.666,1.922){2}{\usebox{\plotpoint}}
\multiput(1222,768)(20.670,1.879){2}{\usebox{\plotpoint}}
\multiput(1266,772)(20.705,1.445){2}{\usebox{\plotpoint}}
\multiput(1309,775)(20.666,1.922){2}{\usebox{\plotpoint}}
\multiput(1352,779)(20.707,1.412){2}{\usebox{\plotpoint}}
\multiput(1396,782)(20.666,1.922){2}{\usebox{\plotpoint}}
\put(1439,786){\usebox{\plotpoint}}
\sbox{\plotpoint}{\rule[-0.600pt]{1.200pt}{1.200pt}}%
\sbox{\plotpoint}{\rule[-0.200pt]{0.400pt}{0.400pt}}%
\put(1279,156){\makebox(0,0)[r]{\scriptsize MM95}}
\sbox{\plotpoint}{\rule[-0.600pt]{1.200pt}{1.200pt}}%
\put(1299.0,156.0){\rule[-0.600pt]{18pt}{1.200pt}}
\put(183,192){\usebox{\plotpoint}}
\multiput(185.24,192.00)(0.500,0.709){78}{\rule{0.121pt}{2.018pt}}
\multiput(180.51,192.00)(44.000,58.811){2}{\rule{1.200pt}{1.009pt}}
\multiput(229.24,255.00)(0.500,0.526){76}{\rule{0.121pt}{1.584pt}}
\multiput(224.51,255.00)(43.000,42.713){2}{\rule{1.200pt}{0.792pt}}
\multiput(270.00,303.24)(0.541,0.500){68}{\rule{1.623pt}{0.121pt}}
\multiput(270.00,298.51)(39.631,39.000){2}{\rule{0.812pt}{1.200pt}}
\multiput(313.00,342.24)(0.604,0.500){60}{\rule{1.774pt}{0.121pt}}
\multiput(313.00,337.51)(39.317,35.000){2}{\rule{0.887pt}{1.200pt}}
\multiput(356.00,377.24)(0.837,0.500){42}{\rule{2.331pt}{0.121pt}}
\multiput(356.00,372.51)(39.162,26.000){2}{\rule{1.165pt}{1.200pt}}
\multiput(400.00,403.24)(1.129,0.501){28}{\rule{3.016pt}{0.121pt}}
\multiput(400.00,398.51)(36.741,19.000){2}{\rule{1.508pt}{1.200pt}}
\multiput(443.00,422.24)(0.927,0.501){36}{\rule{2.543pt}{0.121pt}}
\multiput(443.00,417.51)(37.721,23.000){2}{\rule{1.272pt}{1.200pt}}
\multiput(486.00,445.24)(0.994,0.501){34}{\rule{2.700pt}{0.121pt}}
\multiput(486.00,440.51)(38.396,22.000){2}{\rule{1.350pt}{1.200pt}}
\multiput(530.00,467.24)(0.970,0.501){34}{\rule{2.645pt}{0.121pt}}
\multiput(530.00,462.51)(37.509,22.000){2}{\rule{1.323pt}{1.200pt}}
\multiput(573.00,489.24)(1.268,0.501){24}{\rule{3.335pt}{0.121pt}}
\multiput(573.00,484.51)(36.077,17.000){2}{\rule{1.668pt}{1.200pt}}
\multiput(616.00,506.24)(1.594,0.501){18}{\rule{4.071pt}{0.121pt}}
\multiput(616.00,501.51)(35.550,14.000){2}{\rule{2.036pt}{1.200pt}}
\multiput(660.00,520.24)(1.685,0.501){16}{\rule{4.269pt}{0.121pt}}
\multiput(660.00,515.51)(34.139,13.000){2}{\rule{2.135pt}{1.200pt}}
\multiput(703.00,533.24)(1.685,0.501){16}{\rule{4.269pt}{0.121pt}}
\multiput(703.00,528.51)(34.139,13.000){2}{\rule{2.135pt}{1.200pt}}
\multiput(746.00,546.24)(1.685,0.501){16}{\rule{4.269pt}{0.121pt}}
\multiput(746.00,541.51)(34.139,13.000){2}{\rule{2.135pt}{1.200pt}}
\multiput(789.00,559.24)(1.594,0.501){18}{\rule{4.071pt}{0.121pt}}
\multiput(789.00,554.51)(35.550,14.000){2}{\rule{2.036pt}{1.200pt}}
\multiput(833.00,573.24)(2.259,0.502){10}{\rule{5.460pt}{0.121pt}}
\multiput(833.00,568.51)(31.667,10.000){2}{\rule{2.730pt}{1.200pt}}
\multiput(876.00,583.24)(2.026,0.502){12}{\rule{4.991pt}{0.121pt}}
\multiput(876.00,578.51)(32.641,11.000){2}{\rule{2.495pt}{1.200pt}}
\multiput(919.00,594.24)(2.315,0.502){10}{\rule{5.580pt}{0.121pt}}
\multiput(919.00,589.51)(32.418,10.000){2}{\rule{2.790pt}{1.200pt}}
\sbox{\plotpoint}{\rule[-0.200pt]{0.400pt}{0.400pt}}%
%\put(140.0,82.0){\rule[-0.200pt]{0.400pt}{187.420pt}}
%\put(140.0,82.0){\rule[-0.200pt]{312.929pt}{0.400pt}}
%\put(1439.0,82.0){\rule[-0.200pt]{0.400pt}{187.420pt}}
%\put(140.0,860.0){\rule[-0.200pt]{312.929pt}{0.400pt}}
\end{picture}